\documentclass[conference]{IEEEtran}
\IEEEoverridecommandlockouts
\usepackage{amsmath,amssymb,amsfonts}
\usepackage{algorithmic}
\usepackage{graphicx}
\usepackage{textcomp}
\usepackage{xcolor}
\usepackage[numbers]{natbib}
\usepackage{subfigure}
\usepackage{booktabs} % for professional tables
\usepackage{nicefrac}
\usepackage{hyperref}
\usepackage{comment}
\def\BibTeX{{\rm B\kern-.05em{\sc i\kern-.025em b}\kern-.08em
    T\kern-.1667em\lower.7ex\hbox{E}\kern-.125emX}}

\newcommand{\cready}[1]{{}}

\newcommand{\remove}[1]{}

\newcommand{\zipnn}{\emph{ZipNN}}
\begin{document}

\title{ZipNN: Lossless Compression for AI Models}
\author{
\IEEEauthorblockN{Moshik Hershcovitch$^{1,3}$, Andrew Wood$^4$, Leshem Choshen$^{1,5}$, Guy Girmonsky$^1$, Roy Leibovitz$^6$, Or Ozeri$^1$,\\ Ilias Ennmouri$^2$, Michal Malka$^1$, Peter Chin$^6$, Swaminathan Sundararaman$^1$, Danny Harnik$^1$}
\IEEEauthorblockA{
\textit{IBM Research$^1$, IBM$^2$, Tel Aviv University$^3$, Boston University$^4$, MIT$^5$, Dartmouth College$^6$} \\
moshik1@gmail.com, \{leshem.choshen, Guy.Girmonsky, ilias.ennmouri, Michal.Malka, swami\}@ibm.com, oro@il.ibm.com\\
\{roy.leibovitz.27, Peter.Chin\}@dartmouth.edu, aewood@bu.edu, dannyh@il.ibm.com
}
}

\maketitle

\begin{abstract}
With the growth of model sizes and the scale of their deployment, their sheer size burdens the infrastructure requiring more network and more storage to accommodate these. While there is a vast model compression literature deleting parts of the model weights for faster inference, we investigate a more traditional type of compression -- one that represents the model in a compact form and is coupled with a decompression algorithm that returns it to its original form and size -- namely lossless compression. 

We present \zipnn{}, a lossless compression tailored to neural networks. Somewhat surprisingly, we show that specific lossless compression can gain significant network and storage reduction on popular models, often saving $33\%$ and at times reducing over $50\%$ of the model size. We investigate the source of model compressibility and introduce specialized compression variants tailored for models that further increase the effectiveness of compression. On popular models (e.g. Llama 3) \zipnn{} shows space savings that are over $17\%$ better than vanilla compression while also improving compression and decompression speeds by $62\%$. Using multiple workers and threads, \zipnn{} can achieve decompression speeds of up to 80GB/s and compression speed of up to 13GB/s.   
We estimate that these methods could save over an ExaByte per year of network traffic downloaded from a large model hub like Hugging~Face.  
%\moshik{add the fact that the deocmpression throughput can reach up to 80GB/s}}
\end{abstract}


\section{Introduction}
\label{submission}

With scale, we have learned that models gain stronger abilities and with it, gain popularity.
With scale, models also require more storage space and memory, and with popularity, more communication bandwidth. Taken together, we observe strains on communication bottlenecks that call for efficient solutions. Storage requirements, while often ignored, may accumulate to hundreds or thousands of times the size of a model if checkpoints~\cite{Biderman2023PythiaAS} or distributed updates are to be saved (c.f.,~\ref{sec:related}) \cite{kandpal2023git,
don-yehiya-etal-2023-cold,zhang2021survey}. 

Similarly, models are repeatedly moved around in multiple channels: from a storage hub to inference machines; from training/fine-tuning nodes to the storage backend; between GPU nodes during distributed training, and so on.
Network hubs epitomize the strains by model size. For instance, with over 14.5 GBs and 2.77 M downloads per month from Hugging~Face \cite{Wolf2019HuggingFacesTS} Mistral \cite{jiang2023mistral} alone requires 40 PBs of transferred information a month.

A large body of work has been aimed at reducing model sizes focusing on the parameter sizes and the number of computations in inference. Such methods transform the model into a smaller one in an irreversible fashion. For example, distillation \cite{gou2021knowledge}, pruning  \cite{ma2021effective} and quantization \cite{gholami2021survey_quantization} all lose some information and, potentially, performance, for improved efficiency. Since these methods' main focus is on inference speed, they are bound to the format of an actual running model. As such, they don't necessarily push the space-saving to its limits, and are not stored in the minimal possible way.  %those methods have two unique characteristics, they can hurt performance, and they are restricted to representations that are efficient for compute, not storage (see \S\ref{sec:related}). 

In this work, on the other hand, we follow a more traditional definition of compression typically used for networking and storage. Compression that is also accompanied by a decompression process, returning a model to its origin - namely lossless compression.  
%We review the effectiveness of lossless compression techniques for large models including variations on standard techniques (\S\ref{sec:losseless_res}).%that improve the associated space and network savings.

Surprisingly, we observe (\S\ref{sec:compression}) that even standard lossless compressors like zlib \cite{deutsch1996zlib} or zstd \cite{collet2018zstandard} can achieve %meaningful 
non-negligible savings. While common rationale expects model parameters to have high entropy and therefore be non-compressible, we find that in reality there is less entropy than what the representation offers. Our goal is therefore to find a compression method that maximizes compression benefits for models while also being aware of compression and decompression speeds. 

We identify the source of model compressibility as the floating point range that actually exists in models. 
Specifically, we find that the exponent component in a floating point parameter is highly skewed and therefore highly compressible. 
To this end, we devise a compression method that separates the exponent bits from the rest of the bits that often show no compressibility.  We also identified that the source of compressibility of the exponent is entirely the skewed distribution of single bytes. This means that compressors that search for multi-byte repetitions are both unnecessarily time-consuming and unhelpful in this case, and in particular all of the Lempel-Ziv algorithms \cite{lZ77, LZ78} hardly achieve any data reduction. Instead, we use only {\em Entropy Encoding} and specifically Huffman codes \cite{Huffman1952}, improving both performance and compression ratio in doing so. Another speed improvement comes from identifying the non-compressible parts and avoiding time-consuming compression attempts on them.  
The observations described above are relevant to most of the models, but not all of them. \zipnn{} therefore identifies the characteristics of the model at hand and determines which strategy best suits it. 

The compression benefits are typically tied directly to the model parameter type. The BF16 model family is best compressed, as the exponent in BF16 accounts for half of the total bits (8 of the 16 bits), such models gain approximately 33\% of space savings - reducing the exponent by $\nicefrac{2}{3}$ amounts to reducing the full model size by $\nicefrac{1}{3}$. 
In contrast, typical FP32 models show space saving of just 17\% as the exponent accounts for $\nicefrac{1}{4}$ of the model parameters.   

We classify popular models into categories with distinct compressibility traits. We highlight the \emph{clean models} category, where models have undergone some rounding after the training phase. The rounding allows compression of the other bits (apart from the exponent) and therefore clean models achieve effective compression, at times reaching 55\% savings. For Example, the popular RoBERTa model~\cite{liu2019roberta} falls into this category.  Our compressor is therefore challenged with the task of inline identifying the potential space savings in such a model and compressing it accordingly.  

When comparing \zipnn{} to a state-of-the art compressor like Zstd, we get the following improvements: On BF16 models our compressor achieves a 17\% improvement in compression ratio and a 62\% speedup in compression/decompression. For clean models, this is even larger with a 34\% improvement in space and a 4.6X speedup in compression time and 83\% speedup in decompression. By using multi-threading we show that \zipnn{} can reach a throughput of up to 80GB/s for decompression and up to 13GB/s for compression.   

Finally, we explore the benefits of delta compression and show that by compressing the delta between two similar models one can achieve compression far greater than compressing a standalone model. This is useful for checkpointing and management of model variations.  

The main contributions of this paper can be summarized as:
\begin{itemize}
    \item We observe that lossless compressors may be effective for AI models and identify the source of their compressibility. 
    \item We introduce a \textbf{new lossless compression method} tailored for AI models, which achieves a \textbf{better compression ratio} and \textbf{faster} compression and decompression.
    \item We categorize models according to their compressibility, highlighting that on the popular BF16 models, \zipnn{} can \textbf{reduce the model size by 33\%}, and on so-called clean models by over a half. 
  
    \item We study the potential of delta compression for checkpointing and model versions, learning that although during training all model weights constantly change, fewer bits are changing in every epoch, leading to better delta compression. 

%    \item \moshik{achieve line rate in inference serving system where the model can stay always compress on file sharing}
\end{itemize}
\section{Background}
\subsection{Motivation - use cases}
With small models weighing about a Gigabyte \cite{Devlin2019BERTPO} and large ones Terrabytes \cite{Fedus2021SwitchTS} storage and network overheads become an issue for many purposes. Moreover, common use cases require many model types or model versions and hence increased resources. We list some below as a motivation.
\begin{table}[t]
\caption{Top ranked downloaded models from Hugging~Face.}
\label{table:HF}
\vskip -0.1in
\begin{center}
\begin{small}
%\begin{sc}
\begin{tabular}{lccccc}
\toprule
Model &  Model &  \#Monthly & Rank & Compressed\\
name &  Size & Downloads & & Size\\
\hline \hline
\textbf{Bge} & 0.4 GB & 434M &\#1 & \textbf{42.1\%}
\\
\textbf{Mpnet} & 0.4 %1
 GB & 226M& \#2 & \textbf{82.9\%}
\\ 
\textbf{Bert} &  0.4 GB & 85M & \#3 & \textbf{83.9\%}
\\
\textbf{Qwen} &  3.1 %1
 GB & 50M & \#6 & \textbf{66.9\%}
\\
\textbf{Whisper} & 6.2 GB & 40M & \#7 & \textbf{42.7\%}
\\
\textbf{xlm-RoBERTa} & 2.2 GB & 39M &\#8 & \textbf{42.3\%}
\\
\textbf{Clip} & 0.6 GB & 28M &\#10 & \textbf{49.7\%}
\\
\textbf{Llama 3.1} & 812 GB & 14M &\#20 & \textbf{67.2\%}
\\
\bottomrule
\end{tabular}
%\end{sc}
\end{small}
\end{center}
%\vskip -0.1in
\end{table}

\subsubsection{Model Hubs} Large model repositories or hubs like Hugging~Face \cite{Wolf2019HuggingFacesTS}, Model Zoo \cite{modelZoo}, PyTorch \cite{Pytorch_2019}, Tensorflow \cite{tensorflowHub}, Adapter \cite{adapterhub}, IBM watsonx.data~\cite{watsonxdata} and Qualcomm\textsuperscript{®} AI \cite{QualcommAIHUB} hold a large number of models and serve numerous download requests of popular models. Hugging~Face, the largest of these hubs, revealed in a statement in August 2024, that it holds 1.3M models, with a cumulative storage space of 12PB. They also serve 1 billion daily requests amounting to a network bandwidth of around 6 PetaBytes per day! Table~\ref{table:HF} shows some of the top-ranked models\footnote{Rank as of October 2024.} and their compression ratio (using the methods described in Section~\ref{sec:lossless}). As seen, the potential traffic savings from compression is substantial.% Note that the same trends also apply to models that are not downloaded as often, for Exabmple, the Bloom model offers significant savings due to its large initial size. 

In this use case, there are three ways in which compression can be beneficial, the first is to reduce the amount of data transferred, the second is to reduce the time to download and upload models and the third is to reduce the amount of data stored. In a cloud based hub like Hugging~Face, which runs over the internet, the network benefits are crucial (in particular the first one) (see Section~\ref{sec:e2e}). On the other hand, in use-cases of serving models with high performance, pricey storage, typically the storage savings becomes more important (e.g. see Section~\ref{sec:inference}).

%Compression during distributed training can reduce the networking burden, since the need is not to affect the training, we can use lossless compression. The compression should be fast and by using streaming compression on the CPU with many threads it could be beneficial to reduce the data transferred with a minimal additional latency because of the compression. Regarding GPU compression, future GPU technology are expected to have a direct connection between GPUs. In this case, GPU compression will be the only option since the data will not transfer through the CPU.
% Our testing indicates that there are benefits also in compressing the optimizer and the gradient.

%\moshik{add reference to fast storage use cases the usecase of reduce storage on inference serving system \ref{sec:inference}}

\subsubsection{Distributed and Decentralized Training} During training of large models, training nodes transfer data between them to overcome the need to save the full model and computation on a single GPU/node. In some methods, only the model weights are transferred between nodes, and in other methods, the optimizer weights and gradients are transferred as well \cite{zhao2023pytorch}. Either way, distributed training is usually limited by data transfer between nodes, an issue compression addresses directly. For example: FSDP (Fully Sharded Data Parallel) is a PyTorch method that during training, in addition to transferring model weights, also transfers gradients and optimizer states, causing a networking bottleneck that limits the training of large models.

Another training paradigm is decentralized or federated training, which proposes accommodating training of the same model by different contributors. This ranges from federated learning that contributes gradients \cite{zhang2021survey}, to contributing partially trained models \cite{li2022branch}, from changing the kinds of updates done \cite{lialin2023relora} to relying on volunteer computing \cite{NEURIPS2021_41a60377}, or even relying on different objectives and expertise all merged into the same model \cite{don-yehiya-etal-2023-cold}. All of these methods inherently transfer and store numerous model versions, which also drove dedicated version control frameworks \cite{kandpal2023git}. 

\subsubsection{Checkpoints and Versions} During model creation, multiple intermediate versions of the models are commonly saved. This often includes tests on the training regime such as hyperparameter tuning \cite{Turner2021BayesianOI}. Even during the training of a single model, the current model is periodically checkpointed to recover after a crash, to select the best checkpoint from a few options \cite{Dodge2020FineTuningPL}, for analysis \cite{Biderman2023PythiaAS}, improve performance \cite{JunczysDowmunt2018MarianFN,Sandler2023TrainingTM}, etc. 
Even though saving during checkpointing rarely slows the training time, it does burden the storage (as well as the network), limiting the frequency and amount of saved and shared checkpoints, which are encouraged by the community (e.g.; \cite{Biderman2023PythiaAS,liu2023llm360}). 
%\danny{mention XenHub as prior art for fine tuning}

\vspace{-0.1in}
\begin{figure}[ht]
%\vskip 0.2in
\begin{center}
\centerline{\includegraphics[width=\columnwidth]{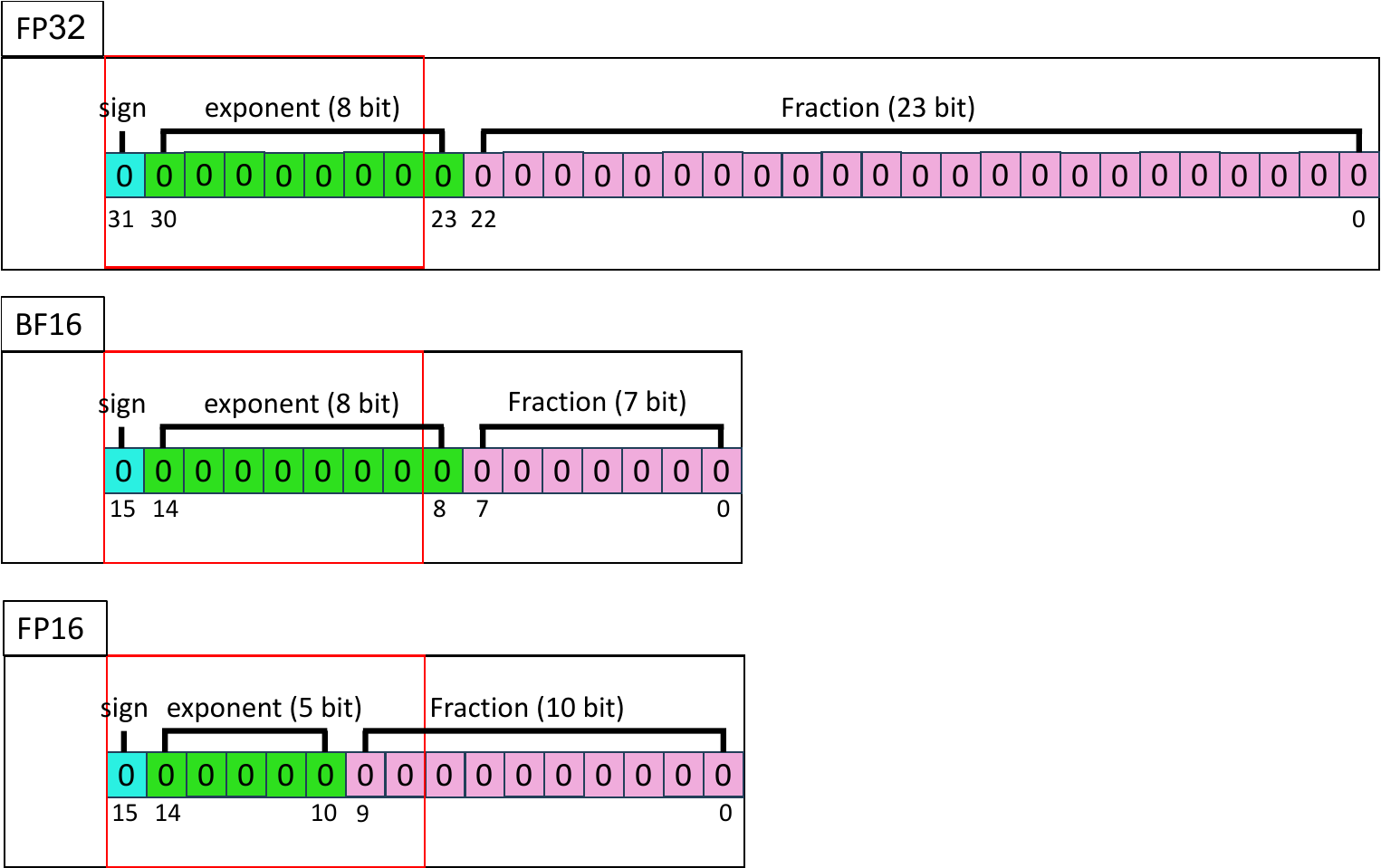}}
%\vspace{-0.25in}
\caption{Representation of FP32, BF16, FP16 floating point types.\looseness=-1 %First byte in red.
%a sign bit + 8 bits exponents + 23 bits of mantissa. , a sign bit + 8 bits exponents + 7 bits of mantissa. FP16, a sign bit + 5 bits exponents + 10 bits of mantissa
}
\label{fig:FP32}
\end{center}
\end{figure}
\vspace{-0.25in}

\subsection{Models Structure and Types}\label{sec:models}  Regardless of the architecture, current models are mainly composed of matrices or tensors of different sizes and a code that can read the parameters in the matrix and convert it to a function. While a layer may contain several such tensors, for brevity we call each tensor a layer. We note that the code is negligible in weight and hence the main focus of our compression is reduced to tensors, or even put more simply, long arrays of numeric parameters.

The type of those numbers is a key factor in the ability to compress the model parameters. 
Parameters typically represent real numbers and as such the most straightforward standard approach is to represent them with {\em floating point} numbers. The floating point format represents a flexible scale with a fixed number of bits which allows for larger numerical ranges. In a nutshell, floating point contains an exponent part - indicating the range in which the real number lies, a {\em mantissa} or {\em fraction} pointing to the actual number within this range, and a sign bit denoting whether a number is positive. For example, FP32 is a 32 bit floating point number with a sign bit, an 8 bit exponent and a 23 bit mantissa (see Figure~\ref{fig:FP32}). 
The real number is calculated by $(-1)^{sign} \cdot 2^{exponent-127} \cdot 1.fraction$. 
Many models are trained with FP32 for its high precision or a mix of precisions. Later during inference, the models sometimes keep their original format, but often choose to forgo some precision for a more compact representation and more efficient computation \citep{shoeybi2019megatron, granite2024granite,dubey2024llama}. 
Perhaps the most popular parameter type used for inference models is BF16~\cite{wang2019bfloat16} which cuts the tail end of the fraction (hence reducing the precision level) but maintains the same exponent as shown in Figure~\ref{fig:FP32}.
Other options include FP16, or quantized models using integer parameters with fixed precision. 

%We found three kinds of models commonly exist, and those have slightly different compression tendencies. One group uses float precision with 32 bytes (FP32), one group saves in FP32, but trained on less, whether implicitly due to optimizations or explicitly saving in a different format than that of training for technical and practical reasons \cite{raffel2020exploring}. The last group saves in a different format, BF32 \cite{wang2019bfloat16}, which gains popularity and as we show in \S\ref{sec:results} yields even better compression rates.

%We did not find any effects of model size, arch (e.g. encoder-decoder) or domain on compressibility, characteristics that commonly affect training and performance \cite{zhai2022scaling}, we hence focus on the effects that seem to matter to compression, such as the groups above.

\subsection{Lossless compression}\label{sec:lossless}
Lossless compressors are the traditional form of compression and are widely used for reducing network and storage overheads in all fields of computing. They consist of two algorithms -- compression and decompression, where after applying both in sequence the output returns to the exact same state. There are countless compression techniques and they vary in the tradeoff between compressibility and compression/decompression time (see for example \cite{squash}).

Throughout the paper, we measure the effectiveness of compression using {\em compression size} in percent. Namely, the percentage of the data that is left after compression -- {\em lower is better}. For example, if the method compresses a GB into a quarter of a GB it has a compressed size of $25\%$.  

The main techniques employed in lossless compression are based on repetition removal (stemming from the seminal works of Lempel and Ziv \cite{lZ77, LZ78} and dubbed LZ compression) and entropy encoding (e.g. \cite{Huffman1952, Riss1976AC}). LZ compressors find multiple-byte repetitions (typically of at least 4 bytes) and replace these with shorter back-pointers, hence saving space. Entropy encoding, on the other hand, looks at the entropy seen in the distribution of single bytes and reduces the length by working at a bit-level granularity, coding popular bytes with short representations. The most popular compressors combine the two techniques (first repetition removal and then entropy encoding), for example Zlib \cite{zlib} and Zstd~\cite{zstd}.\footnote{In our experiments, we chose Zstd as the underlying compressor/decompressor due to its superior speed vs.\ compression tradeoff \cite{zstd, squash}.} A second family of compressors, that favors speed over compression ratio, relies solely on LZ repetition removal, e.g. LZ4 or Snappy \cite{lz4, snappy}.   

\section{Compression for Models}\label{sec:compression}

In this section, we introduce \zipnn{} -- lossless compression specifically tailored for model compression.  
We will first discuss compression of the majority of models and model contents. 
As mentioned in the introduction, we classify models into two categories - regular models and what we call {\em clean} models. Clean models are models that have undergone various techniques like number rounding or transformation between parameter types. These transformations typically leave many bits as zeros and increase model compressibility. 
However, once such a model is re-trained or fine tuned, it quickly becomes less compressible and behaves like a regular model. 
Regular models form the majority of models, and are models that were trained and remained unmodified after the training phase.   
In addition, we note that model formats also contain some metadata and at times a few layers that behave differently (for example tensors of parameters that contain integers). These typically constitute a negligible part of the model and hardly affect the model compression ratio. 
Later in the section, we will discuss clean models and variations. 

% We further present ZSTD compression found to be most promising among the methods tested and present Byte Grouping, a technique adapted for language models that further improves the effectiveness of compression. 
%Note that in this paper, we evaluated only compression run in the CPU, since the GPU computations and GPU memory are more valuable resources, whereas CPU computations and CPU memory are typically abundant. 

\subsection{Regular Model Compressibility}\label{sec:model_compressibility}
Simply deploying standard compressors to a model produces mixed results. For most models, using an LZ-type compressor like LZ4 or Snappy yields no gains at all. This is expected since it requires the data to contain sequences of bytes that repeat. However, model tensors are both noisy and unstructured - meaning that parameters typically do not have an affinity with their neighbors. This makes repetitions that span multiple parameters scarce. 

Using a compressor that also combines entropy coding, such as Zlib or Zstd, does show some compressibility. Initially, one may expect models to be non-compressible and show high entropy, as parameters may encode unpredictable information and differ from each other. This is correct to a certain degree, but in reality, the actual range in which parameters reside is typically limited, which reduces the entropy and opens the door for compression to be effective. 

\begin{figure}[ht]
%\vskip 0.2in
\begin{center}
%[trim={left bottom right top},clip]
\centerline{\includegraphics[width=\columnwidth, trim={18mm 35mm 12mm 35mm},clip]{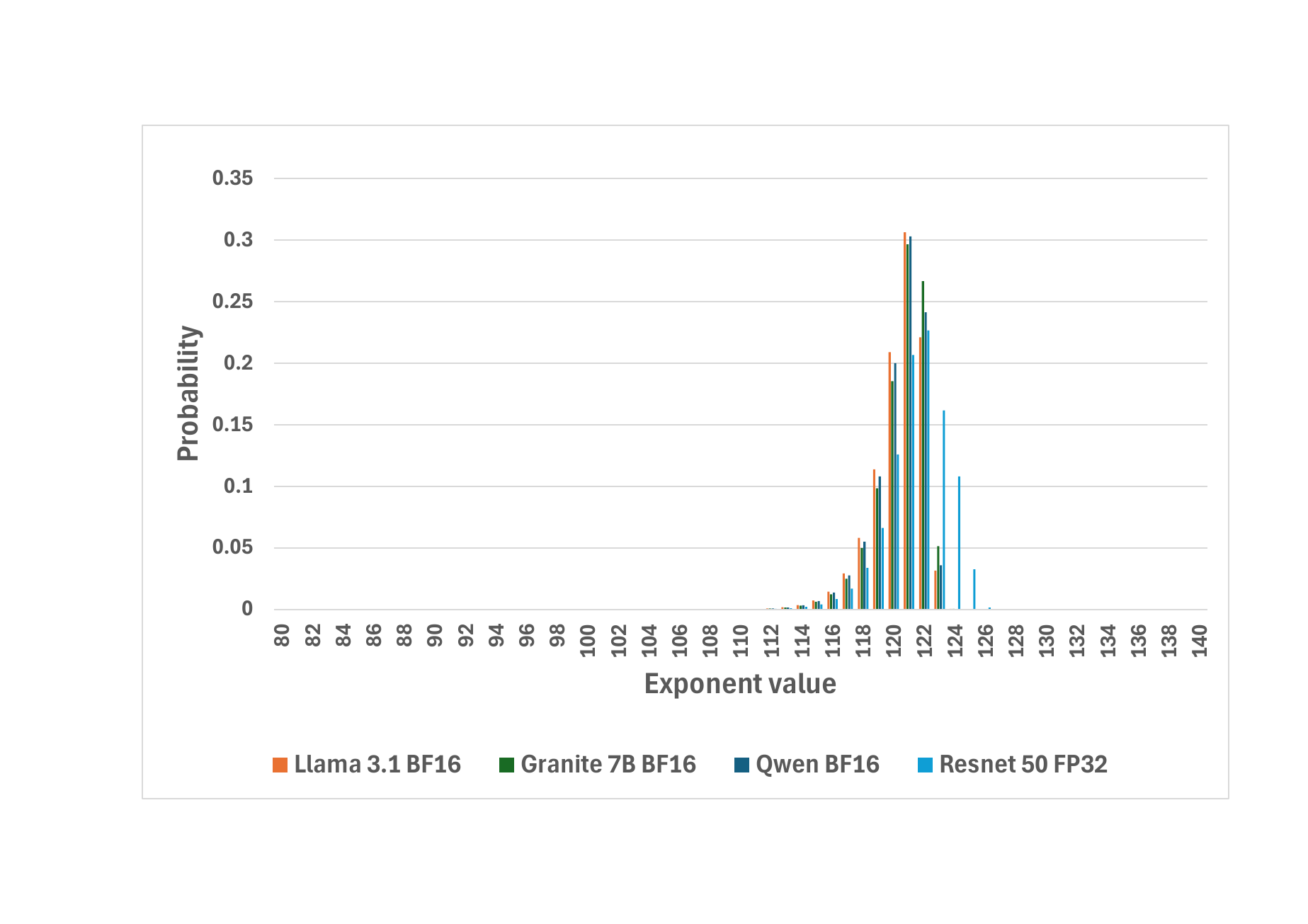}}
%\vspace{-0.25in}
\caption{{\bf Histogram of exponent values} for four different models. The graph is focused only on middle values and no exponent values appear outside this range. Based on 1GB taken from the middle of the model. }
\label{fig:exponent}
\end{center}
\end{figure}

A deeper dive shows that models indeed contain a high level of randomness in their parameters, but this is true for the fraction and sign bit parts of the floating point parameter. The exponent on the other hand is highly skewed and is the source of the compressibility. Figure~\ref{fig:exponent} shows the distribution of the exponent in four different models. This distribution is highly skewed, and strikingly similar between different models, whether BF16 or FP32. It is very similar among language models, but also close when testing other model types, such as Resnet, which is an image model. Out of the 256 possible values in the exponent, we see only around 40 that actually appear (50 in the image model). Moreover, the top 12 values account for almost 99.9\% of all parameters (17 in the image model).

\vspace{-0.1in}

This highly skewed distribution can be explained as an artifact of the way models are trained. 
Model weights are initially set in the space of [-1,+1] and the training process scarcely pushes weights out of this range, perhaps since weights can have sufficient impact without the need to grow substantially. This explains why exponent values do not tend to exceed 128, which translates to the range [-1,1]. On the other hand, we see that weights are also not found in the lower ranges of extremely high precision. This is likely a result of various choices in the training process, for instance, optimizers (e.g. the Adam optimizer) add some noise in order to avoid division by zero. This noise is set in the order of $2^{-23}$ (corresponding to an exponent of around 99), and thus weights do not fall much below this noise level.

\begin{figure}[ht]
\vskip 0.2in
\begin{center}
\centerline{\includegraphics[width=\columnwidth, trim={40mm 16mm 15mm 40mm},clip]{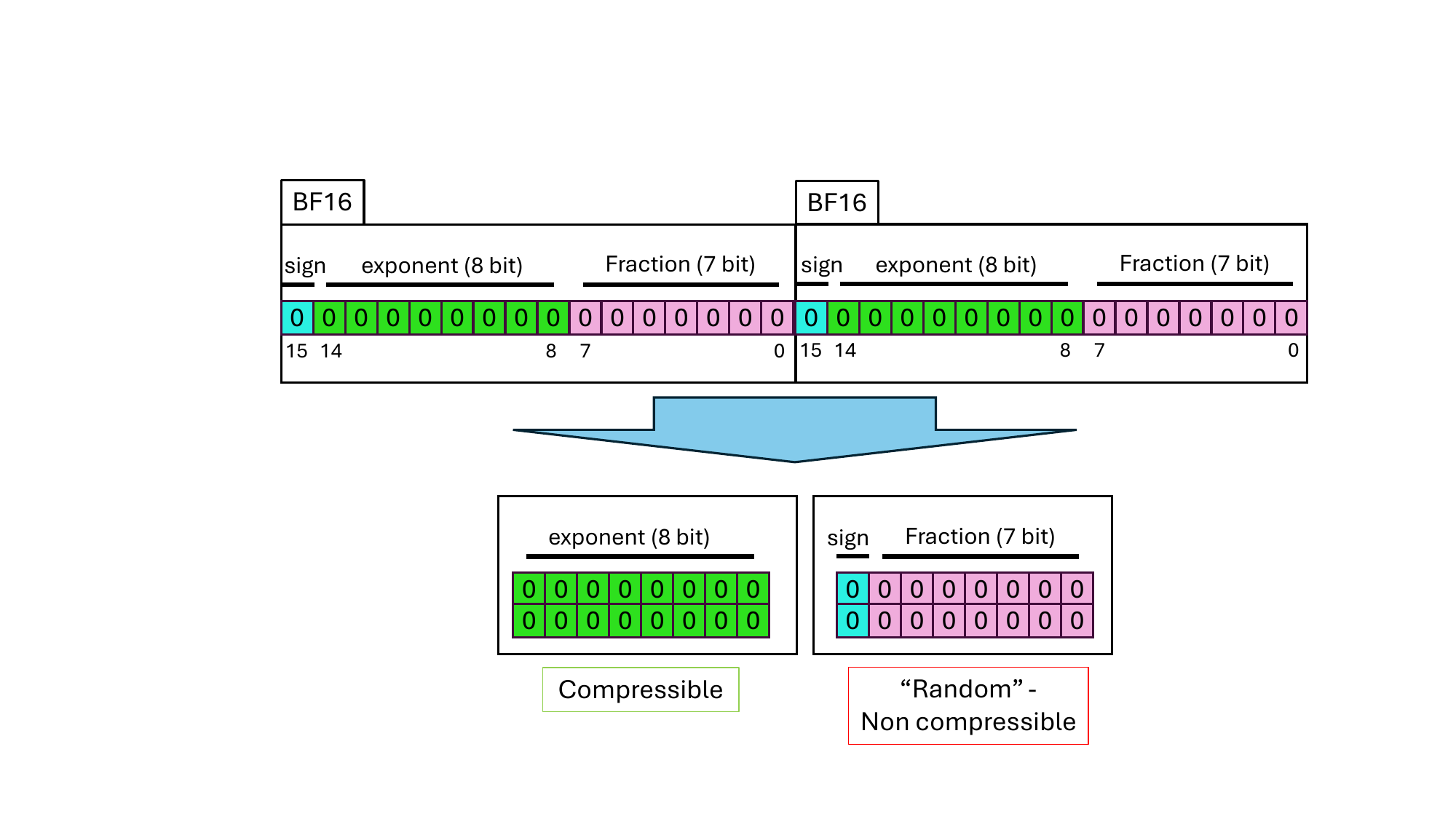}}
\vspace{-0.1in}
\caption{Exponent Extraction for BF16 - group all exponents into a separate compression stream.}
\label{fig:BF16}
\end{center}
\end{figure}

\paragraph{Exponent Extraction.}
No matter the reason that the exponent is skewed, we leverage this skewed distribution for tailoring our compression method. 
It is clear that mixing exponent data with either the sign bit or the fraction bits will interfere with the compression of the exponent. We therefore choose to rearrange the data in a way that separates the exponent data. We call this {\em exponent extraction} and it is 
depicted in Figure~\ref{fig:BF16} for a BF16 based model. 

\vspace{-0.25in}

\vspace{-0.15in}

\paragraph{Huffman only Compression.}
In addition, we observe that there is no structure in the tensors, namely that the exponents are not ordered in any way. Whatever repetitions are found by an LZ type compression, these are most likely ''random" in the sense that they are an artifact of the skewed distribution. We therefore forgo the LZ compression part and use only an entropy encoder and specifically we use Huffman encoding.
This proved to be helpful both in terms of compression/decompression performance, which is expected, but also in improving the compression ratio. It turns out that the random repetitions found by the LZ phase are naturally short, and not only do they not save much, but they also interfere with the effectiveness of the Huffman encoder (by adding back-pointers to the mix). 
In order to ensure that the repetitions found are indeed ''random", we shuffled the parameters in a model randomly and then compressed the exponents using Zstd. The shuffled version reached nearly the same compression ratio (up to 0.05\%).
This experiment shows that Huffman encoding provides a very good trade-off of time vs.\ space. An FSE entropy encoder achieves a slightly better compression ratio (0-2\%) at the expense of significant performance penalty (at times over 2X).% \danny{Moshik, do we have numbers?}   

\begin{figure}[ht]
%\vskip 0.2in
\begin{center}
\centerline{\includegraphics[width=0.95\columnwidth, trim={0mm 0mm 0mm 0mm},clip]{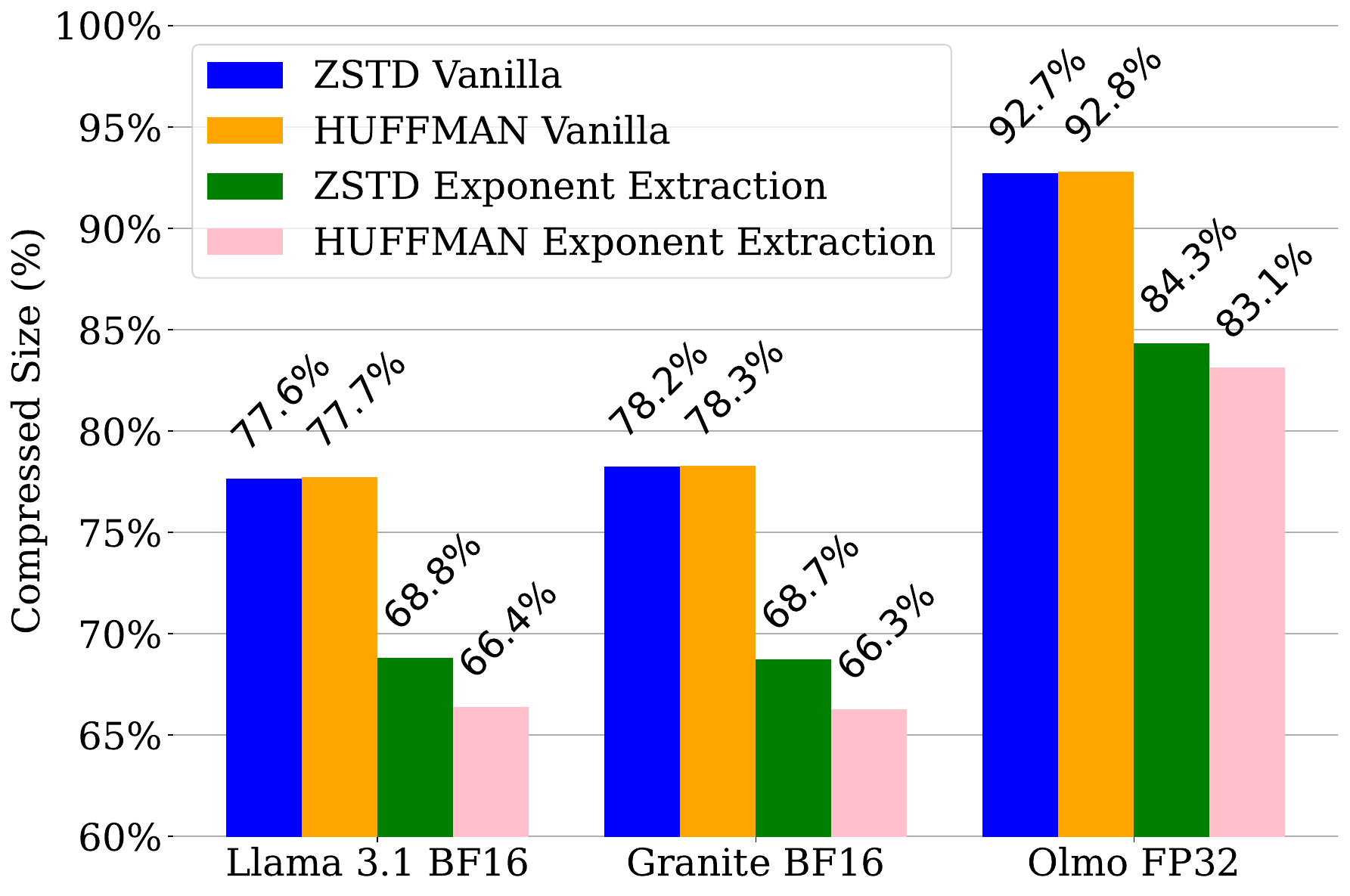}}
%\vspace{-0.25in}
\caption{Breakdown of the contributions of Exponent Extraction and Huffman only encoding to compression ratio.}
\label{fig:BF16_ablation}
\end{center}
\end{figure}

Figure~\ref{fig:BF16_ablation} shows the compression ratio benefits of each of our steps. We see that using Huffman without exponent extraction is only helpful for speed over Zstd. However, once we divide the exponent from the rest of the data, using Huffman vs Zstd also achieves better compression (as well as speed).

\vspace{-0.15in}

When compressing regular models we found that the exponent compresses by approximately a 3X factor (namely compressed size of 33\%) whereas the fraction and sign bits hardly compress at all. The overall savings is thus dictated by the weight of the exponent in the parameter. That is, for BF16 the compressed size is $\frac12\cdot 33.3 + \frac12 \cdot 100 \approx 66.6$, whereas for FP32 it is $\frac14\cdot 33.3 + \frac34 \cdot 100 \approx 83.3$. As a result, the main benefit of compressing regular models lies in BF16 models (which fortunately are most of the new models used in practice). 
We show the effect on performance of the exponent extraction and Huffman only encoding in Section~\ref{sec:speed}

\subsection{Compressing Clean Models}

As mentioned earlier, most models only show compressibility in the exponent. However, we also see some models that are surprisingly more compressible, and show compressibility also in the fraction part of the parameters. We call these clean models, since after further fine tuning they typically lose their extra compressibility. That being said, these include some of the most popular models, such as Bge, RoBERTa, Xlm-RoBERTa, Whisper, and Clip. 

Clean models pose some additional challenges in devising the best possible compression method for models. While for regular models one can assume the fraction is incompressible and not compress it, for general models we need to identify that the fraction is indeed compressible and furthermore, choose the correct compression method for the fraction part. 
We first describe our observations regarding how to best compress clean models. 

\begin{figure}[ht]
%\vskip 0.2in
\begin{center}
\centerline{\includegraphics[width=\columnwidth, trim={0mm 0mm 20mm 110mm},clip]{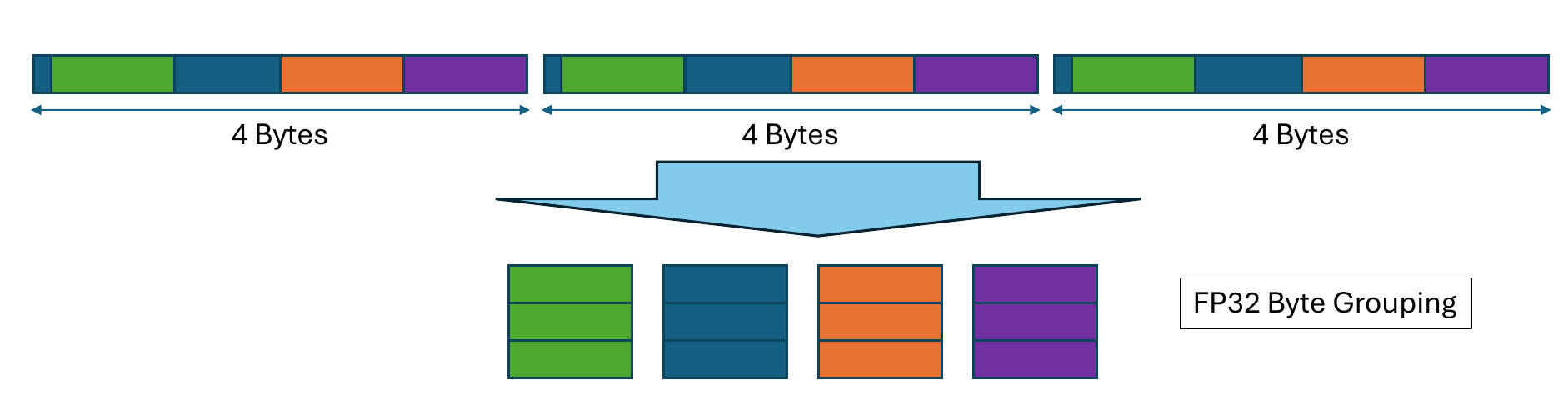}}
%\vspace{-0.25in}
\caption{An example of Byte Grouping for FP32. The first group consists of the exponents. The rest of the data is split into 3 groups, one per each byte in the parameter.}
\label{fig:BG}
\end{center}
\end{figure}

\paragraph{Byte Grouping.}
When dealing with FP32 models, the first observation regarding fraction compression is that also within the fraction different bytes have different compressibility. For example, if rounding is used, then the least bits of the fraction will be zeros while the most bits will remain more or less random. 
This suggests that the different bytes of the fraction should also be separated into different compression streams. We call this technique {\em byte grouping} and it is depicted in Figure~\ref{fig:BG}.  
\vspace{-0.15in}

\begin{figure}[ht]
%\vskip 0.2in
\begin{center}
%[trim={left bottom right top},clip]
\centerline{\includegraphics[width=\columnwidth]{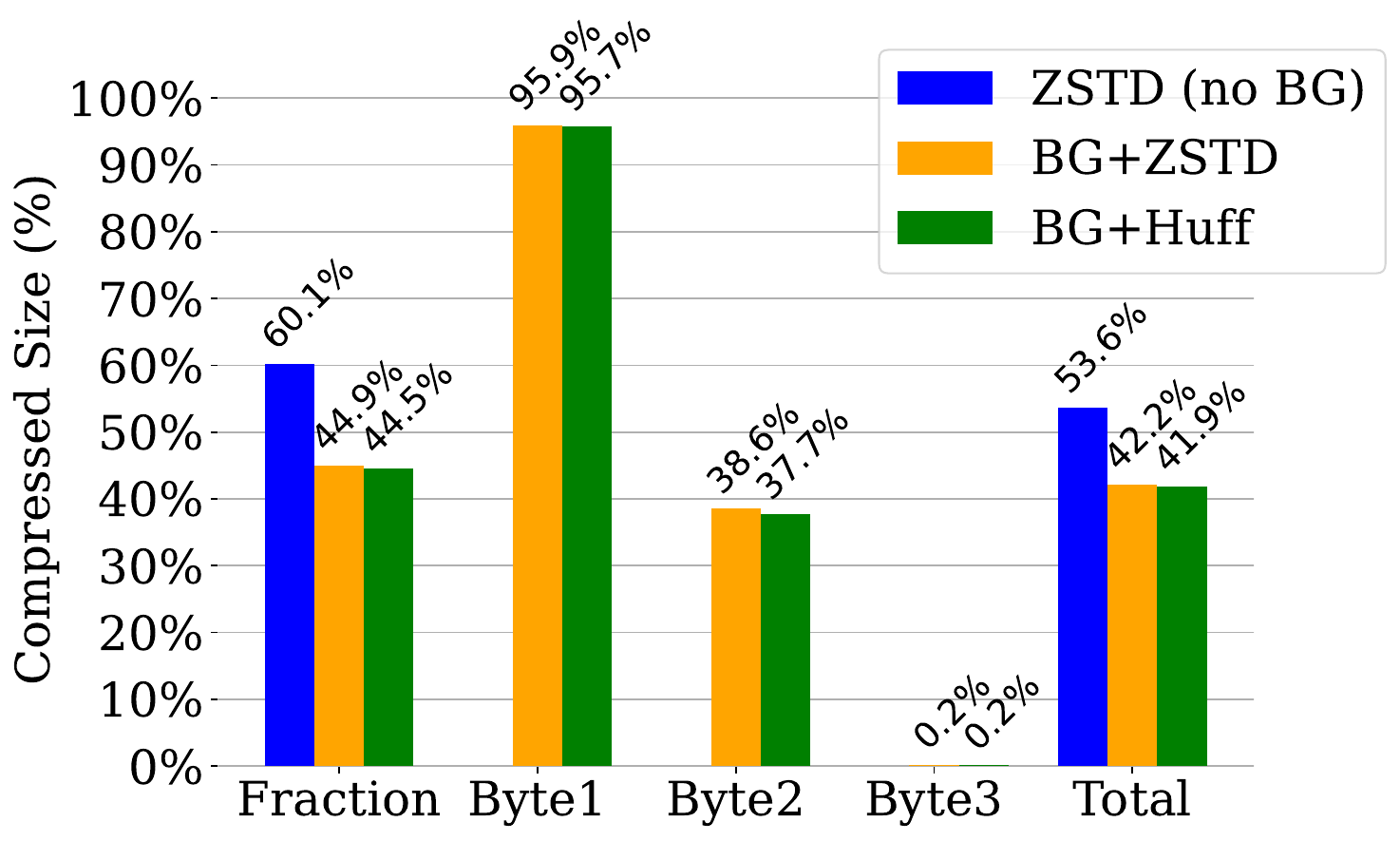}}
%\vspace{-0.25in}
\caption{Compression ratios for the clean model xlm-RoBERTa (FP32) with and without byte grouping (BG), including a breakdown of compressibility of the 3 fraction bytes. Total includes the fraction and exponent parts.}
\label{fig:clean}
\end{center}
\end{figure}

Like the exponent, Huffman only decoding wins out for the byte groups that are compressible except for all zero streams that can simply be truncated and replaced by a header (the Huffman implementation that we use does this automatically).
Figure~\ref{fig:clean} shows an example of the effect of compressing a clean model (xlm-RoBERTa), with and without byte grouping.
When the fraction is broken down into bytes we see different behavior for the different byte groups. Byte 1 is barely compressible, while byte 3, being all zeros, is highly compressed. Byte 2 is quite compressible and compresses a little better with Huffman only.   

\vspace{-0.15in}

\begin{table*}[t]
%\vskip -0.15in

\caption{Compressed size for various Models using \zipnn{} with Byte Grouping. Based on compressing 1GB from the middle of a model (for large models), and the entire model, excluding the first 10MB (headers etc...) for smaller models.}
%\vskip -0.2in
\begin{center}
\begin{small}
\begin{sc}
\begin{tabular}{lcccccr}
\toprule
Model & Param & Model & Compressed  &  Breakdown\\
name & Type & Size &  Size & to Byte Group  \\
\hline 
\hline
\textbf{Falcon-7b} & BF16 & 14.4 %3
 GB& \textbf{66.4\%} & (32.8\%, 100\%) %8.75\% 
)
\\
\textbf{Bloom} & BF16 & 328.2 
 GB&  \textbf{67.4\%} & (34.8\%, 100\%) %8.6\% 
 \\
\textbf{openllama-3B} & BF16 & 6.9 
 GB & \textbf{66.4\% } & (32.7\%, 100\%) %10.1\%
\\ 
\textbf{Mistral} & BF16 & 14.5 
 GB & \textbf{66.3\% } & (32.5\%, 100\%) %10.1\%
\\ 
\textbf{Llama-3.1} & BF16 &  
 16 GB & \textbf{66.4\% } & (32.8\%, 99.9\%) %10.1\%
\\ 
\hline
\textbf{Wav2vec} & FP32 & 1.2 GB &  \textbf{83.3\%} & (33.0\%, 100\%, 100\%, 100\%) 
\\
\textbf{Bert} & FP32 & 0.4 GB &  \textbf{83.0\%} &  (32.6\%, 99.5\%, 100\%, 100\%) %7.6\%
\\ 
\textbf{Olmo} & FP32 & 5.1 GB  & \textbf{83.1\%} &  (32.5\%, 100\%, 100\%, 100\%) %7\%
\\
 \textbf{stable-video-diffusion} & FP16 &
 4.27 GB & \textbf{84.8\%} & (69.6\%, 100\%) 
 \\
\textbf{CapybaraHermes-Mistral} & FP16 & 14.5 GB & \textbf{84.4\%} & (68.8\%, 100\%)  
)
\\
\hline
\textbf{XLM-RoBERTa} & FP32 & 1.1 GB &  \textbf{41.8\%} &  (33.9\%, 95.6\%, 37.5\%, 0.0\%) %27\%
 &
\\

\textbf{Clip} & FP32 & 1.7 GB & \textbf{48.1\%} &  (33.1\%, 100\%, 45.9\%, 13.4\%) %19.5\%

 \\ 
\textbf{T5 base} & FP32 & 0.8 %3
 GB & \textbf{33.7\%} & (34.6\%, 100\%, 0.0\%, 0.0\%) %22.1\%
\\
\textbf{llama2-13B} & FP16 &  26 GB & \textbf{66.6\% } & ( 64.2\%, 69.0\%) %10.1\%
\\ 
\textbf{tulu-7B} & FP16 & 13.5 
 GB & \textbf{66.6\% } & (64.2\%, 68.9\%) %10.1\%
\\ 

\hline
\bottomrule
\end{tabular}
\end{sc}
\end{small}
\end{center}
%\vskip -0.1in

\label{tab:Summary}
\end{table*}

\paragraph{Identifying compressibility.}
Identifying incompressible data in advance has the benefit of speeding up compression time. Various methods have been discussed in prior work to achieve this, e.g.~\cite{tozip2013}. We propose a simple method that builds on the fact that within each byte group compressibility seems to be consistent. We compress a chunk, and if it is not compressible we skip the compressibility of the following few chunks before trying to compress again (the specific number varies with the actual implementation).   
The rationale in retrying to compress is to identify a possible change in behavior between different layers or parts of the model. %\roy{What are we trying to say here? - rephrased...}

%The second observation is that with long streams of zeros or any long stream of constant bytes, repetition removal is far more effective than Huffman encoding. This is because Huffman encoding, by design cannot compress more than 8X (replacing a byte by 1 bit). So often, in clean models it is advisable to use a combined compressor like Zstd for some of the fraction bits. 
%For this purpose we design an auto detection mechanism to decide between three options: compress with Huffman only, compress with Zstd or do not compress. 
\subsection{Compressibility summary}
Table~\ref{tab:Summary} presents the compressibility of various models using \zipnn{} with a breakdown of compressibility of the various byte groups. The first byte group consists of the exponent. We see all BF16 models achieve a space saving of about $\frac13$ of the model size. Regular models with parameter types FP32 or FP16 show less impressive space savings whereas clean models of these type show much better space savings. In clean FP32 models byte grouping plays a key role as seen in the breakdown. We also see clean models in the FP16 family that are likely the result of transformation from BF16 models (which has a shorter fraction part than FP16).

\begin{figure*}[ht]
%\vskip 0.2in
\begin{center}
\subfigure[]{\includegraphics[width=0.32\textwidth]{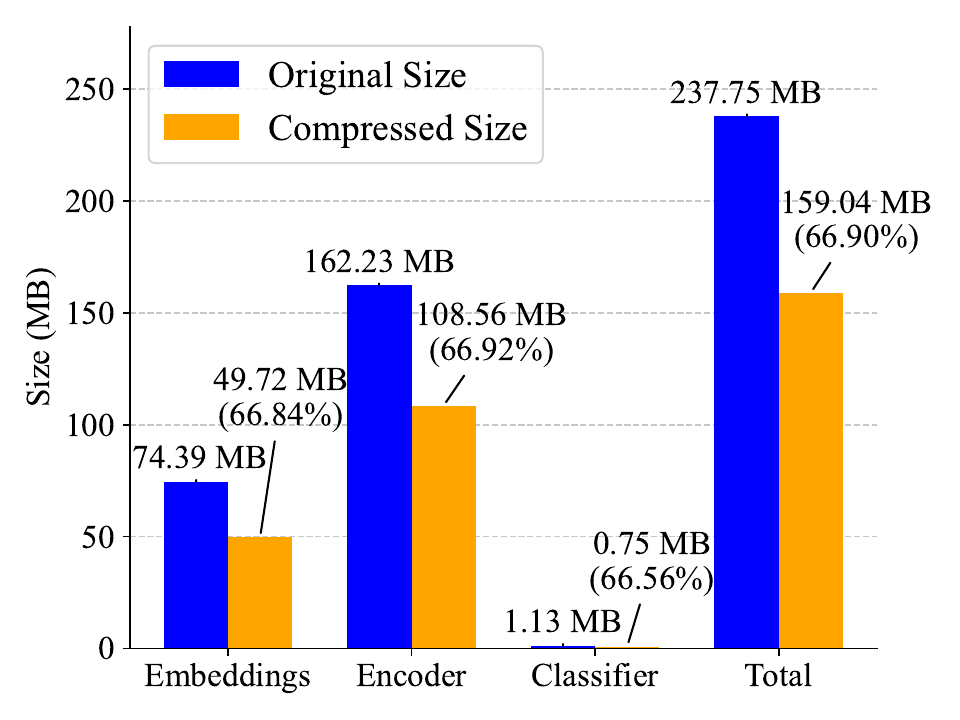}}
\subfigure[]{\includegraphics[width=0.32\textwidth]{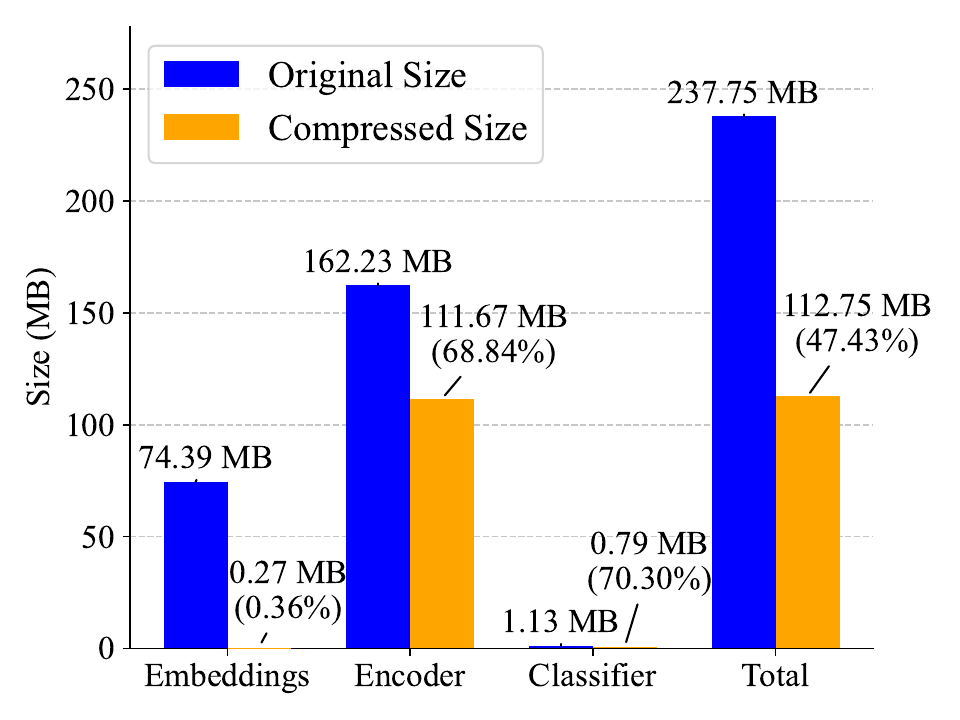}}
\subfigure[]{\includegraphics[width=0.32\textwidth]{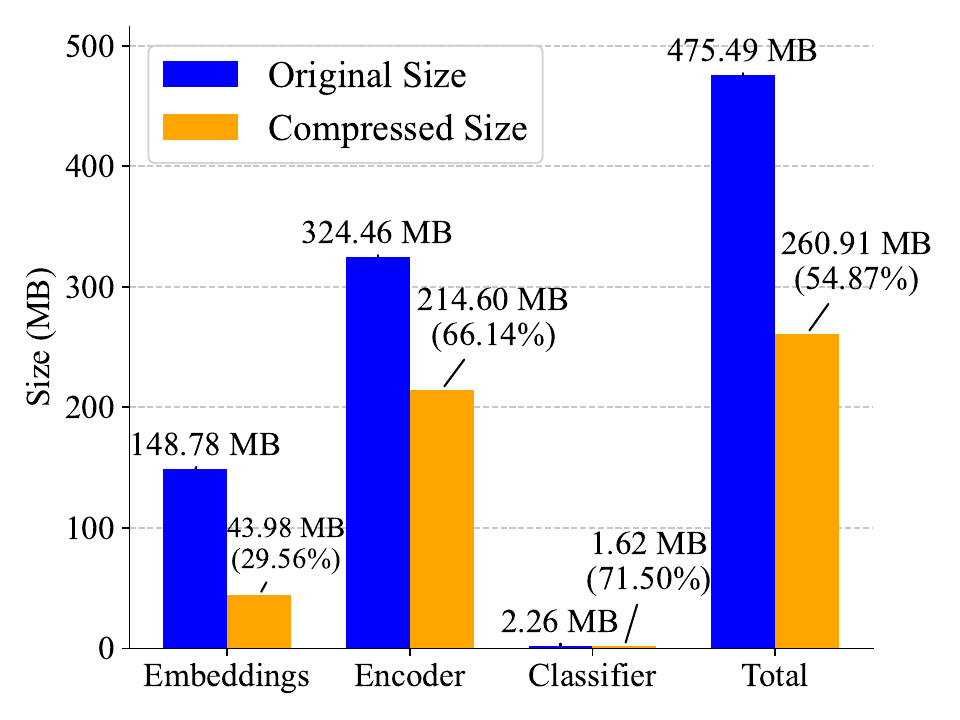}}
%\vspace{-0.25in}
\caption{Compressibility of the various layers in the Roberta {\bf Model} (a) , {\bf Gradients} (b)  and {\bf Optimizer} (c) during training. Compression for all layers uses exponent extraction. The embedding layer in the gradients and optimizer is then compressed using Zstd while all other layers use Huffman.}
\label{fig:gradients}
\end{center}
\end{figure*}
%\vspace{-0.15in}

\section{Beyond Model compression}\label{sec:beyond}

So far we focused only on compression of full models in standalone form. In this section, we discuss how our methods fare when compressing artifacts of the training process, including Optimizer, Gradient and checkpointing data. 

\subsection{Gradients and Optimizers}\label{sec:gradients}
Gradients and Optimizers are derivatives of the training process. These artifacts are required in order to continue the training process and as such take up substantial network bandwidth (in the case of distributed training) or storage space (in the case of checkpoints).
Gradients and Optimizers are often of equal in size to the models \cite{anil2020scalable} and therefore can also benefit from compression. 

We investigate a BF16 version of RoBERTa under finetuning and generate intermediate gradients and optimizers, as well as the model itself. Interestingly, we found that gradients and optimizers compress better than the actual model. While the model compressed size is, as expected, around 66\%, the optimizer is at 54\% and gradient at 47\%. The source of the extra compressibility is found in the different layers of the model. Figure~\ref{fig:gradients} shows a breakdown of the compressibility in the various model layers. Interestingly, the token embeddings layer is extremely compressible in the gradients and optimizers, whereas in the model itself, this layer is not different than other layers. The general layers in optimizers and gradient compress to around $66\%$, slightly better than these layers in the model itself. In addition, we find that both in gradients and optimizers, the embedding layer is compressed significantly better with Zstd (unlike the model itself).

\begin{figure*}[ht]
%\vskip 0.2in
\begin{center}
\subfigure[]{\includegraphics[width=0.32\textwidth]{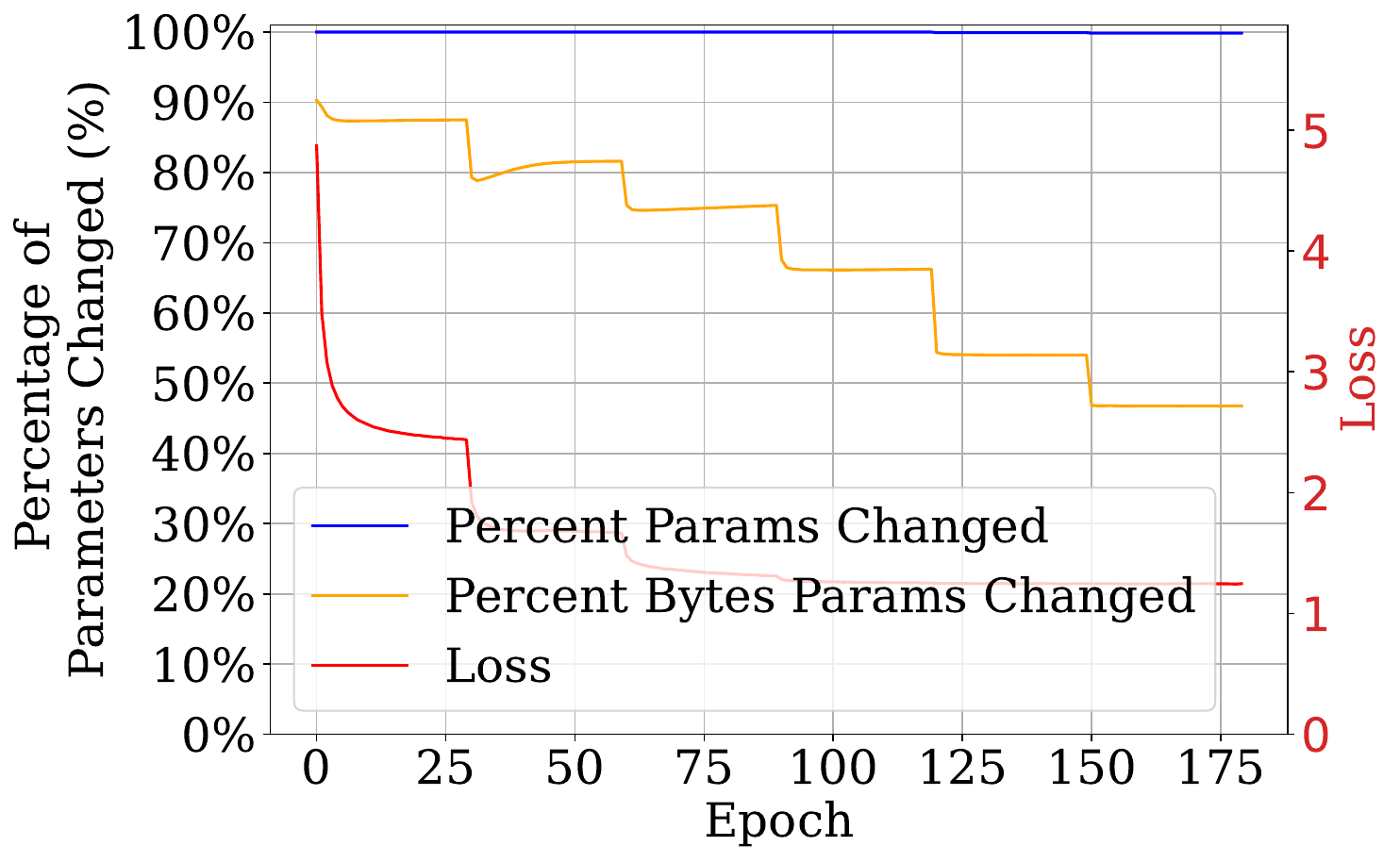}}
\subfigure[]{\includegraphics[width=0.32\textwidth]{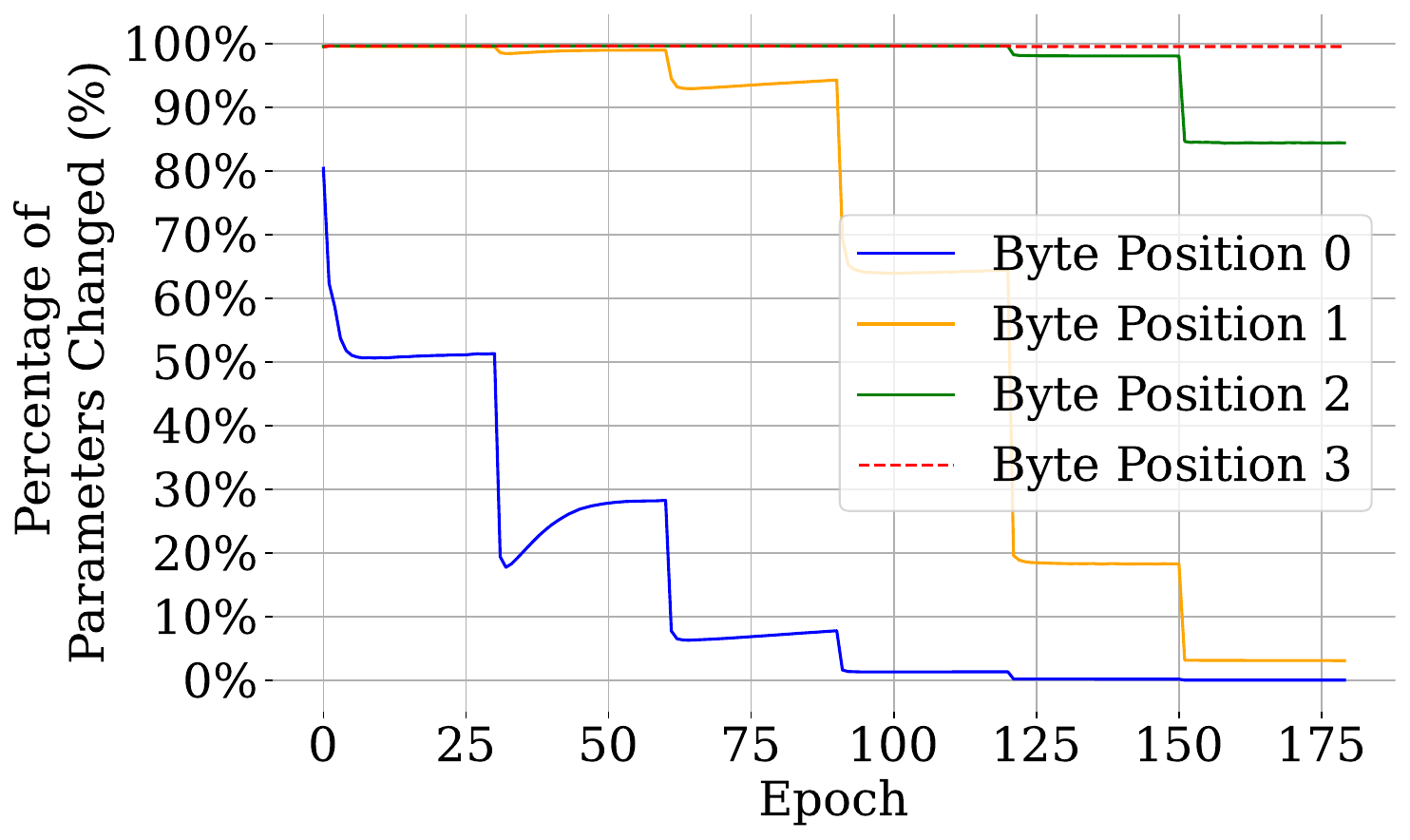}}
\subfigure[]{\includegraphics[width=0.32\textwidth]{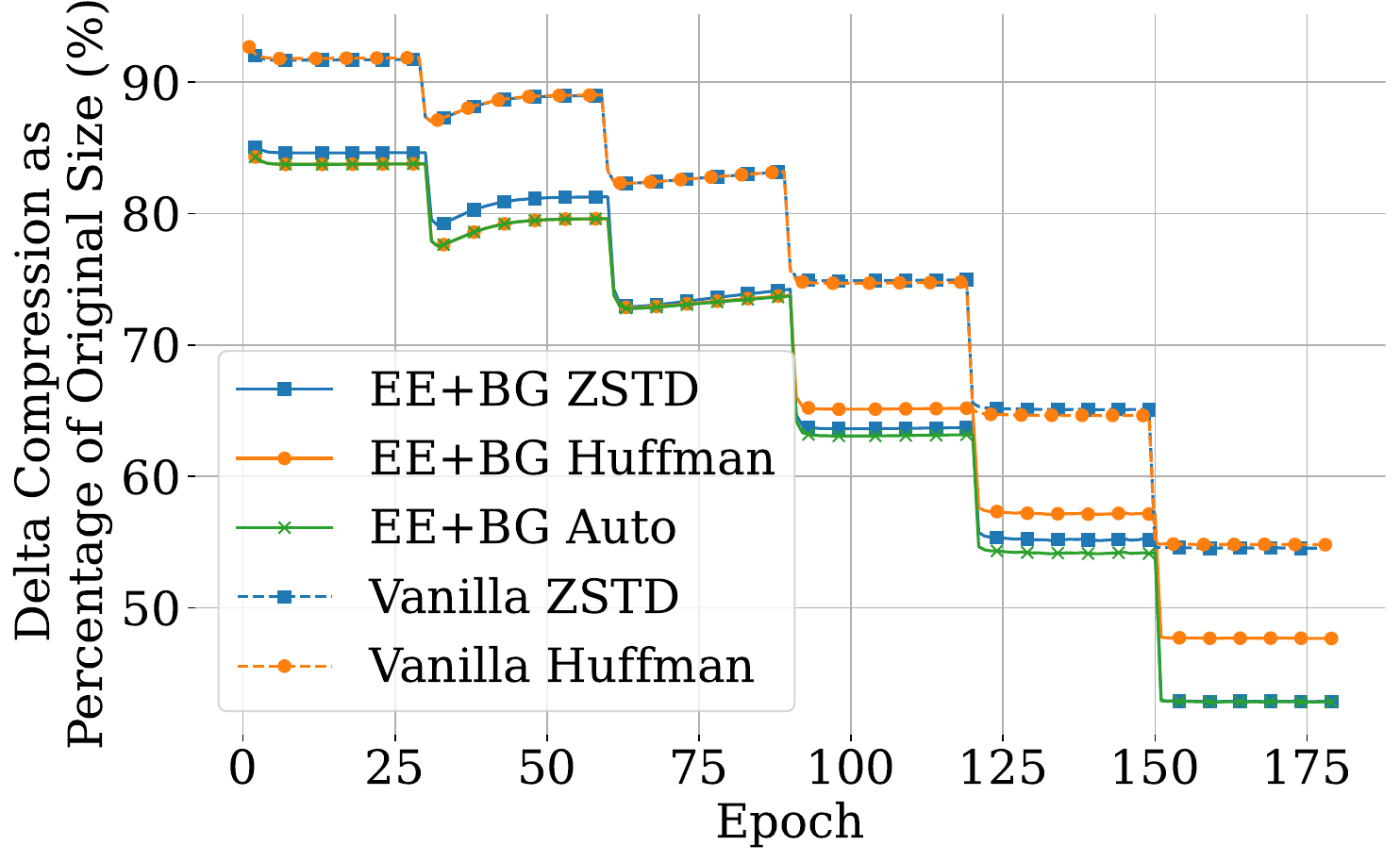}}
%\vspace{-0.25in}
\caption{Resnet 18 finetuning test with checkpoints taken every epoch. (a) shows the amount of change in parameters and bytes as a function of epoch; (b) breaks down this change according to byte groups; (c) shows the delta compression effectiveness with various techniques. The steps seen in the graphs coincide with the steps of the learning rate scheduler. } 
\label{fig:resnet}
\end{center}
\end{figure*}
%\vspace{-0.15in}

\begin{figure*}[ht]
%\vskip 0.2in
\begin{center}
\subfigure[]{\includegraphics[width=0.32\textwidth]{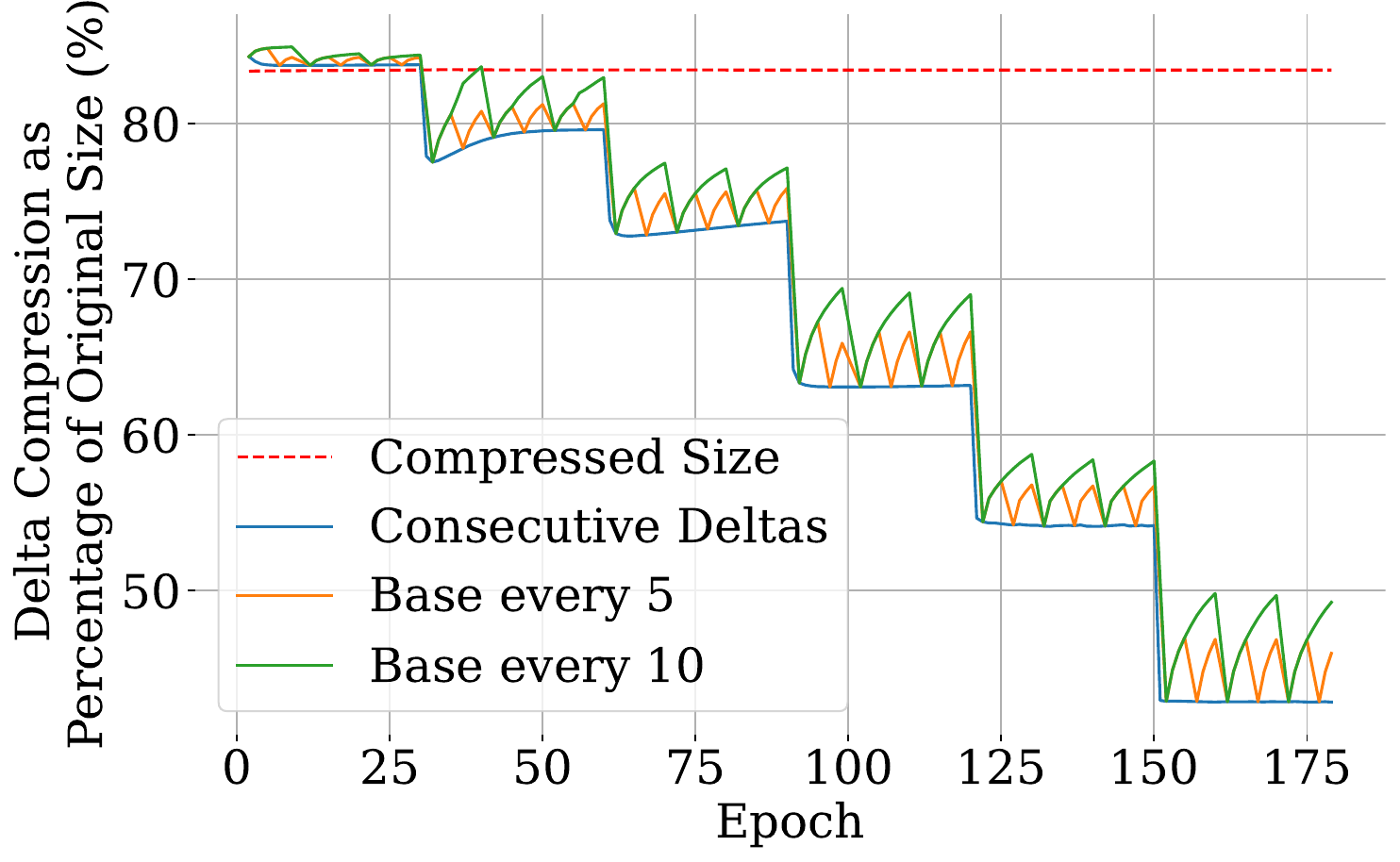}}
\subfigure[]{\includegraphics[width=0.32\textwidth]{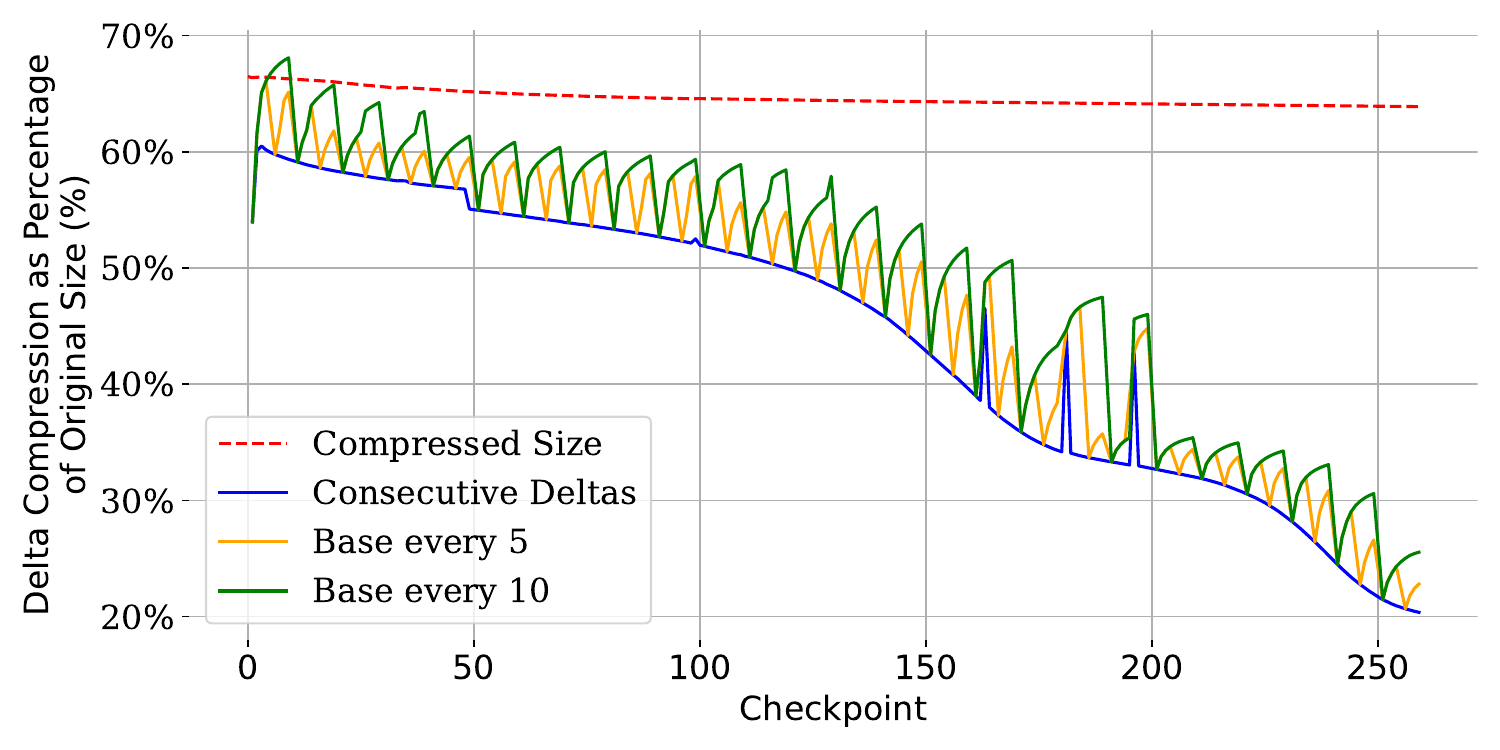}}
\subfigure[]{\includegraphics[width=0.32\textwidth]{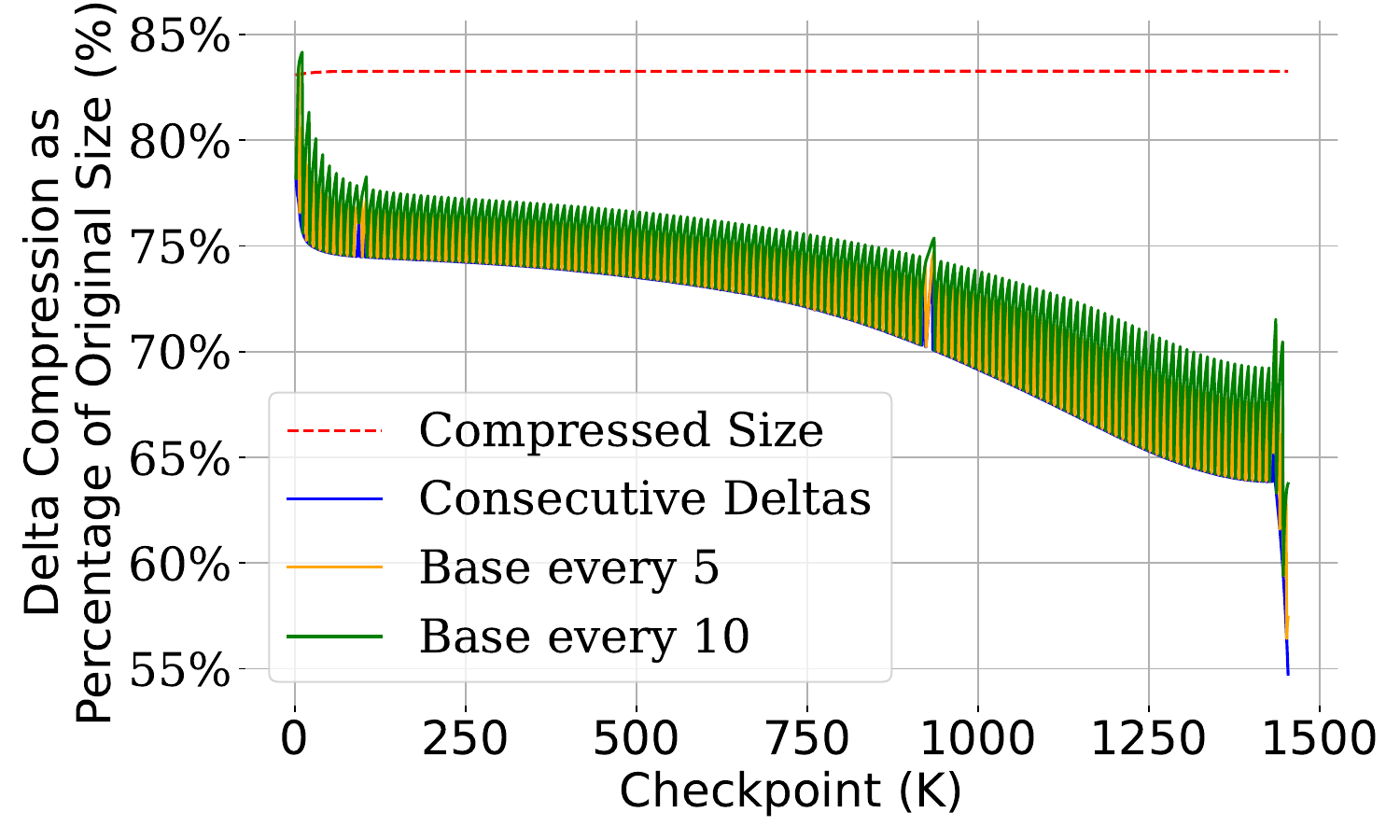}}
%\vspace{-0.25in}
\caption{Compression of checkpoints with periodic bases for (a) ResNet (FP32); (b) Amber (BF16); and (c) Olmo (FP32). In the graphs, we ignore the space of the periodic full bases.} 
\label{fig:bases}
\end{center}
\end{figure*}
%\vspace{-0.15in}

\subsection{Checkpoints and Delta Compression}\label{sec:deltas}

When models have high similarity, one strategy to optimize storage and network transfer is to save a base model and for the rest of the models only store the differences from this base model \cite{kandpal2023git}. We refer to compressing those differences as \emph{delta compression}. To reconstruct a model, one only needs to apply the delta to the base model. A straightforward approach to delta compression is to compute the difference between the two models (e.g. using XOR or subtraction) and compress this delta using a lossless compressor. In our work, we used XOR for the delta as it is easily reversible and does not require any extra bits. 
A natural use case in which delta compression proves useful is checkpointing. 
In checkpointing, we repeatedly store models that have limited change between them. Fine tuning often changes models in small quantities. Hence the delta between the results of consecutive training epochs has the potential to be more compressible than stand-alone models. Note that the main benefit in this use case is storage space rather than offloading from the GPU. Offloading from GPU can, for the most part, be done offline, and we want to refrain from spending extra GPU cycles for compression at the GPU. Hence we focus on CPU based compression for this task. 
We studied this while finetuning a Resnet 18 model (FP32). Figure~\ref{fig:resnet}(a) shows the amount of change between consecutive checkpoints as the process progresses and the loss function drops. We see that while all parameters in the model change in each epoch, when broken down to bytes, more and more bytes remain unchanged as the training converges. Figure~\ref{fig:resnet}(b) breaks this down according to byte groups. Here we see that the exponent byte has the least changes, whereas the least bits in the fraction have the most change. All byte groups show a growing number of steady bytes as the training converges. 
This suggests that byte grouping is helpful for delta compression as well, as our tests in  Figure~\ref{fig:resnet}(c) corroborate.

\paragraph{Auto Detection of Compression method.} While investigating compressibility of deltas, we realized that at some point it is worthwhile to compress the delta using Zstd over Huffman. This is mostly a function of the number of zeros in the delta (unchanged bytes). For this purpose we devised an auto-selection mechanism that decides between Huffamn and Zstd based on two criteria. For each chunk, \zipnn{} counts the number of zeros in the chunk, as well as the length of the longest zero sequence. By running a simulation we found that Zstd compresses better than Huffman if the number of zeros exceeds 90\%. It also outperforms Huffman if there is any long sequence of zeros (which may happen if certain layers are ruled as non-modifiable by the training mechanism) and we set the detection level at 3\% of the chunk size. Figure~\ref{fig:resnet}(c) also compares running with Huffman, vs.\ Zstd vs.\ the Auto detection method. We see that Huffman outperforms Zstd in the first two steps of the learning rate scheduler, and this flips after the $3^{rd}$ step. Auto manages to pick correctly and is always at least as good as the better method.  

\paragraph{Periodic Base.}
A major drawback of delta compression is that in order to recover a checkpoint one must obtain both the base and the delta. But if the checkpoints are all stored using delta compression, this would result in long chains of deltas, making the recovery process prohibitively expensive. 
Instead, it is customary to periodically store a full checkpoint (compressed standalone), serving as a base for the next $k$ checkpoints. 
Suppose that we store a base every 10 checkpoints, then the longest chain of deltas would be of length 9. 
Another approach is to store a periodical base and always do the delta with respect to the last full base. This means that there are never delta chains, but rather only pairs of base and delta. On the other hand, if the period is for example 10, we may be doing a delta against a base that is 9 epochs ago, and may achieve worst delta compression. 
In Figure~\ref{fig:bases} we study what the compression would be if we use a periodic base for a period of 5 and 10. We use the self trained ResNet18 as well as two models for which training checkpoints have been made public (Amber and Olmo). 
Depending on the model and the epoch we see that using a base at distance 5 or even 10, while not as optimal as consecutive deltas, is still far better than standalone compression.

Another use-case is when a hub or user stores multiple models with high similarity (regardless of checkpointing). One source for such occurrence is when multiple models are trained or fine tuned from the same base model. For example, we found in Hugging~Face 3 variations of RoBERTa trained on tweets which are tuned for different purposes - detecting irony, detecting offensive language, and detecting abuse. %(For the exact model names see Appendix~\ref{app:deltas_models}). 
As standalone models, their compressed size is 83.7\% on average. However, compressing the delta of each of the pairs achieves a compressed size of 56\% on average.

%This approach was shown to be effective for checkpointing models during training. \danny{How do we present this? Originally it referred to the unpublished paper }
%We note that it is also effective when storing fine tuned models stemming from a base model, a practice which is quite common.  

\section{Implementation and Evaluation}\label{sec:eval}
\subsection{Implementation}\label{sec:implementation}
We implemented \zipnn{} and it is now an open source project~\cite{ZipNNRepo}. 
It's core is written in C (2000 lines of code) and the wrappers, scripts and tests are all written in Python (4000 lines of code). 
We use the Zstd v1.5.6 library as well as Zstd's underlying Huffman implementation as the underlying compressors.  
For our tests in Python, we use zstandard 0.23.0, torch 2.4.0, and numpy 2.1.0.

\noindent {\bf Chunking.} 
When designing \zipnn{}, we aimed to create an implementation that could process small chunks independently, allowing each chunk to be handled in parallel. This architecture is ideal for GPUs, which contain many cores that, while less powerful than CPU cores, excel in concurrent tasks. %For instance, the NVIDIA H100 has over 14,000 CUDA cores.
Our implementation includes two levels of granularity: \textbf{chunk level} and \textbf{byte-group level} within each chunk. By default, each chunk is of size 256KB. In the BF16 format byte-groups are sized at 128KB whereas in FP32 byte-groups are of size 64KB.
Parallelism can be conducted at the chunk level and the byte-group compression/decompression processes. 
Auto decisions on compression method are done at a byte-group granularity.   

\noindent {\bf Metadata and parallelism.} 
During compression, due to the fixed-sized compression chunks, one can easily parallelize the compression process to multiple workers.  
However, during decompression, the chunks are of variable size. To enable parallel processing, we add a map for the whole model containing metadata for each byte-group and each chunk.

\noindent {\bf Hugging Face Integration.}
For smooth integration with Hugging Face's Transformers library \cite{wolf-etal-2020-transformers}, we implemented a mechanism that automatically decompresses downloaded models, reorders symbolic links in the local cache, updates metadata, and removes the compressed files. We also included an option for manual compression or decompression of models in the local cache.

\subsection{Compression and Decompression Speed - Single Thread}\label{sec:speed}
  %using multiple cores.
Table~\ref{tab:regular} shows the speed benefits of \zipnn{} vs.\ vanilla compressors on 3 representative models. Two regular models (BF16 and FP32) and one clean model (FP32). Zstd was used with default configuration. \zipnn{} consists of Exponent-Extraction and Huffman only compression while EE-Zstd denotes Exponent-Extraction with Zstd compression on the exponent. This was based on 10 runs over 1GB from the middle of the model and the maximum standard deviation observed was 2\%.
The tests were run on an Apple M1 Max machine with 10 cores and 64GB of RAM running macOS Sonoma 14.3 using a single thread and on a single core.

The exponent-extraction and byte grouping carry a performance penalty,  
%that is amplified by the fact that the compressor works harder if it actually manages to find some repetitions, 
yet the use of Huffman only encoding makes up for this and we manage to improve both on speed and on compression ratio simultaneously. 
The variation in speeds between the various models is also explained by their compressibility. While in the regular FP32 model $\frac34$ of the model is non compressible and mostly skipped, in the BF16 this accounts for only $\frac12$ of the model and in the clean model even less than that.  Note that we tested LZ4 and Snappy on these models and while they are faster than all methods, they gain zero compression savings.

\begin{table}
\caption{Comparing speeds of compression methods.}
\vskip -0.1in
    \centering
    \begin{tabular}{clccc}
    \toprule
     Model    & Comp.  &  Comp. & Comp.& Decomp.  \\
     name  &  method  &  size & speed & speed  \\
           &   &  (\%) & (GB/Sec) & (GB/Sec)  \\
    \hline
    \hline
       Llama-3.1  & Zstd & 77.7 & 0.71 & 1.02 \\
        BF16 & EE+Zstd & 68.8 & 0.51 & 1.21\\
         & \zipnn{} & \textbf{66.4} & \textbf{1.15} & \textbf{1.65} \\
        % our+FSE & 66.2 & 0.79 & 0.8 \\
    \hline
        Olmo-1b  & Zstd & 92.3 & 0.97 & 1.02 \\
        FP32 & EE+Zstd & 84.4 & 0.82 & 1.97\\
         & \zipnn{} & \textbf{83.2} & \textbf{1.64} & \textbf{2.48} \\
         % our+FSE & 83.1 & 1.22 & 1.45 \\
         
    \hline
        xlm-  & Zstd & 57.4 & 0.18 & 0.77 \\
        RoBERTa & EE+Zstd & 46.7 & 0.42 & 0.89 \\
        FP32 & \zipnn{} & \textbf{42.9} & \textbf{0.83} & \textbf{1.41} \\
         % our+FSE & 41.5 & 0.44 & 0.47 \\
        
    \bottomrule
    \end{tabular}
    \label{tab:regular}
\end{table}

\begin{figure*}[ht]
%\vskip 0.2in
\begin{center}
\subfigure[]{\includegraphics[width=0.32\textwidth]{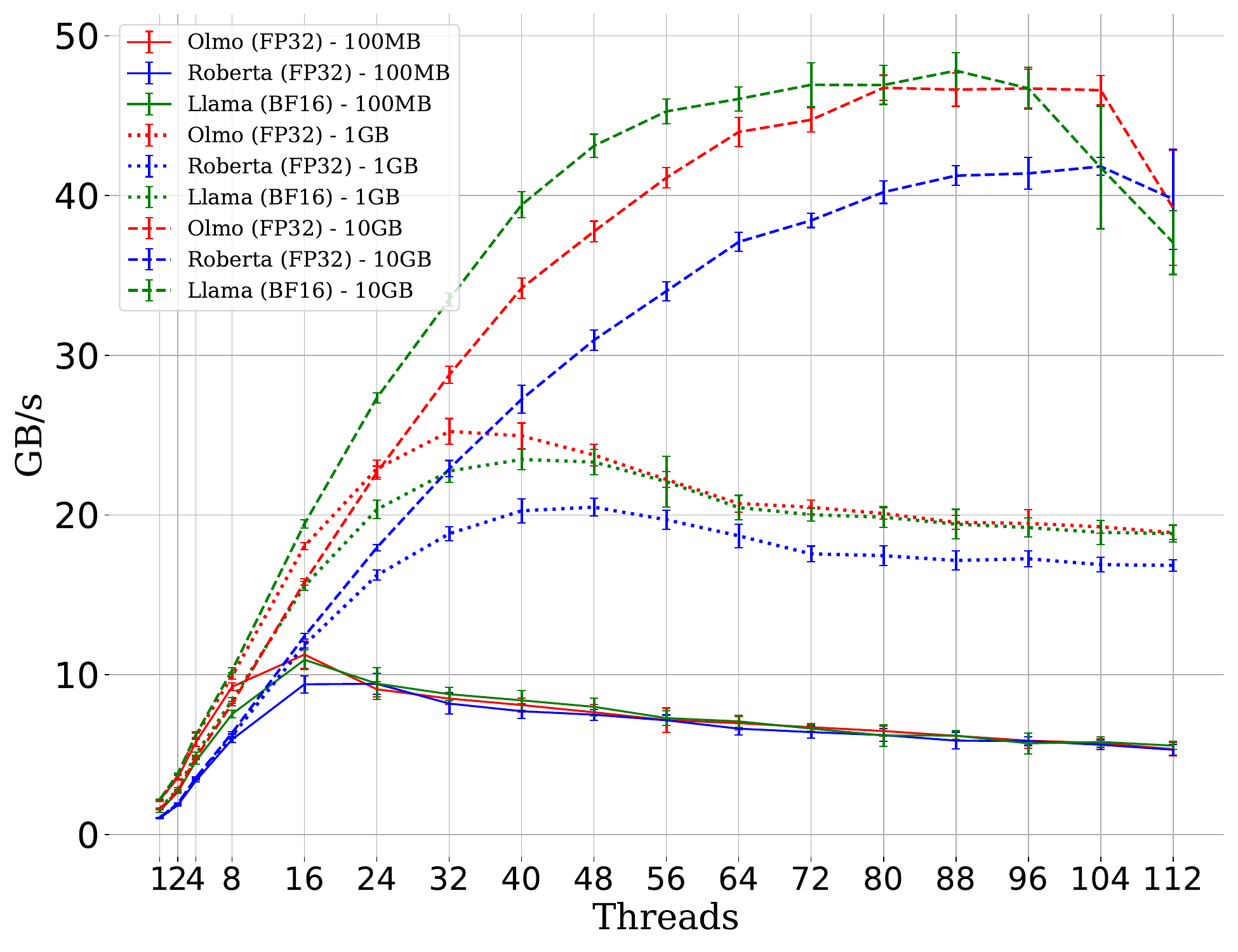}}
\subfigure[]{\includegraphics[width=0.32\textwidth]{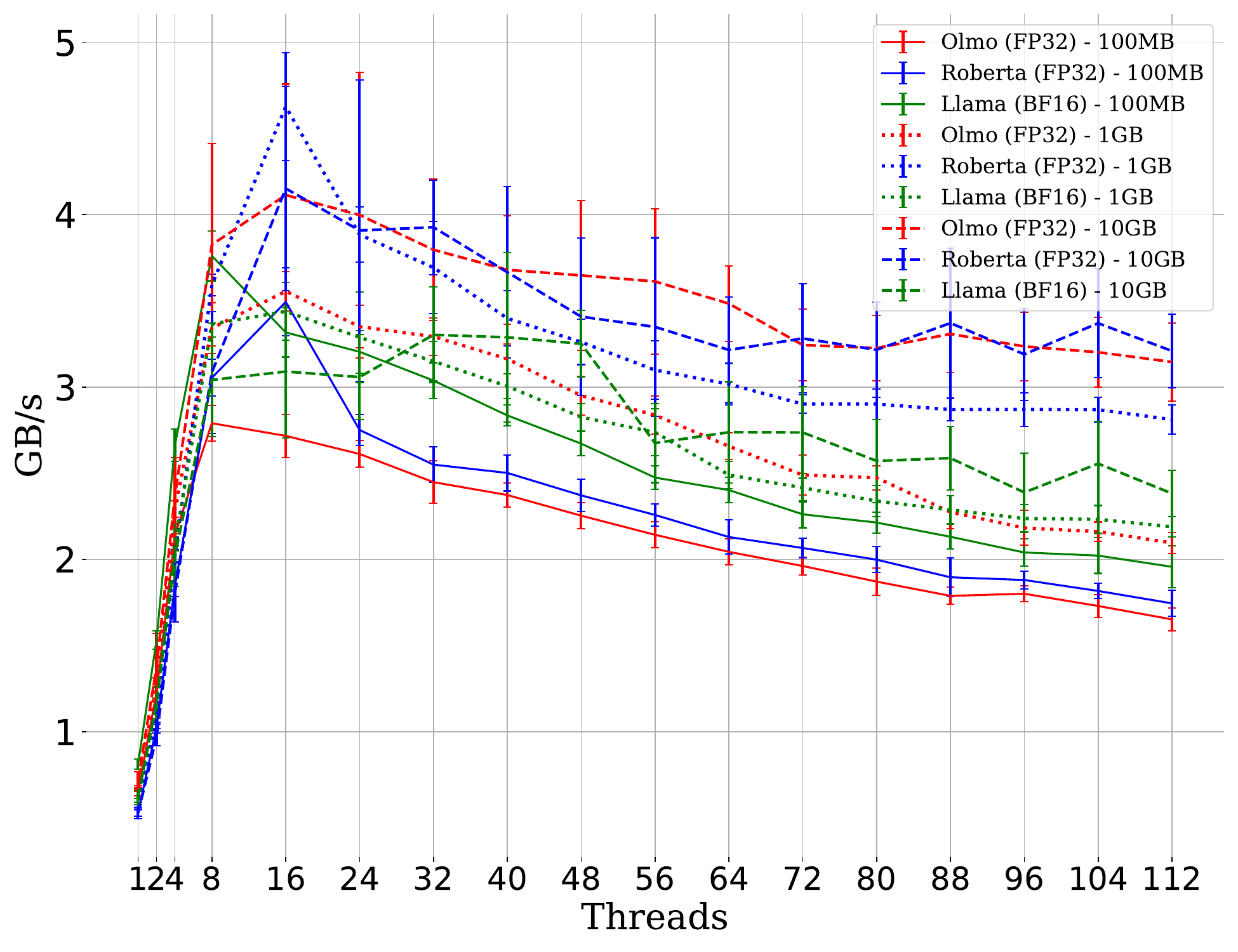}}
\subfigure[]{\includegraphics[width=0.32\textwidth]{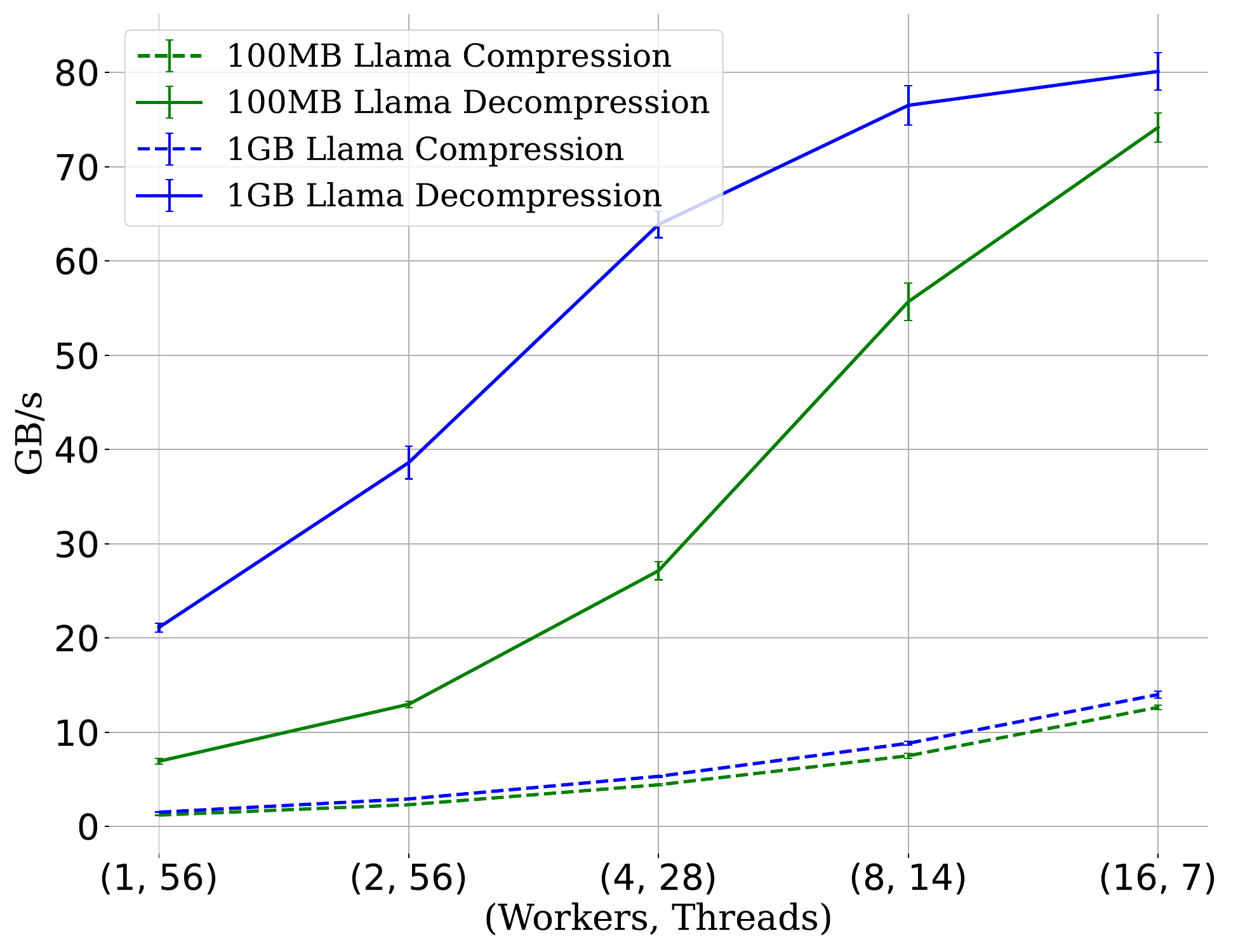}}
%\vspace{-0.25in}
\caption{Decompression throughput (a) and compression throughput (b) as a function of the number of threads used for compressing 100MB, 1GB, and 10GB across three  models. (c) focuses on a Llama 3.1 while also varying the number of workers. The block size here is the size per worker. } 
\label{fig:multi}
\end{center}
\end{figure*}
%\vspace{-0.15in}

\subsection{Multi-threading}\label{sec:multithreading}

We also implemented a multi-threaded version and tested it on a high-performance OpenShift cluster, utilizing a pod equipped with Intel Xeon Platinum 8480+ processors. The CPU features two NUMA nodes with a total of 224 cores (56 cores per socket, 2 sockets per NUMA node) and 2TB of DRAM. 
Figures~\ref{fig:multi}(a) and~\ref{fig:multi}(b) show the results of a single worker as a function of number of threads. We see that the size of the model is critical in utilizing many cores and decompressing 10GB reaches much better throughput (more than 45GBs) than compressing 1GB, let alone 100MB.
The number of threads to reach peak decompression speed varies accordingly. The compression throughput is much lower and peaks at around 16 threads. %The decompression results reveal that Roberta ('clean model'), which decompresses the entire data size, achieves lower decompression throughput compared to Olmo (FP32), which decompresses only $\frac{1}{4}$  of the data, and Llama, which decompresses $\frac{1}{2}$ of the data size. We note that compression demonstrates significantly lower throughput and doesn't scale as effectively as decompression.

We further evaluated \zipnn{} with multiple workers to avoid running across NUMA nodes. Each NUMA node contains 56 threads (corresponding to all threads from a single socket). This multi-worker approach scales better than using a single worker as seen in Figure~\ref{fig:multi}(c), reaching up to 80GB/s for decompression throughput and up to 13GB/s for compression with 16 workers and doing so while each worker gets a block size of as low as 100MB.

%Results were obtained from an internal system ("shit2")—this reference will be updated - Unsure what you meant here Moshik?

% I wanted the figured to be one below another...

\remove{

\begin{figure}[ht]
%\vskip - 0.1in
    \begin{minipage}[b]{0.99\linewidth}
        \centering
        \includegraphics[width=\linewidth]{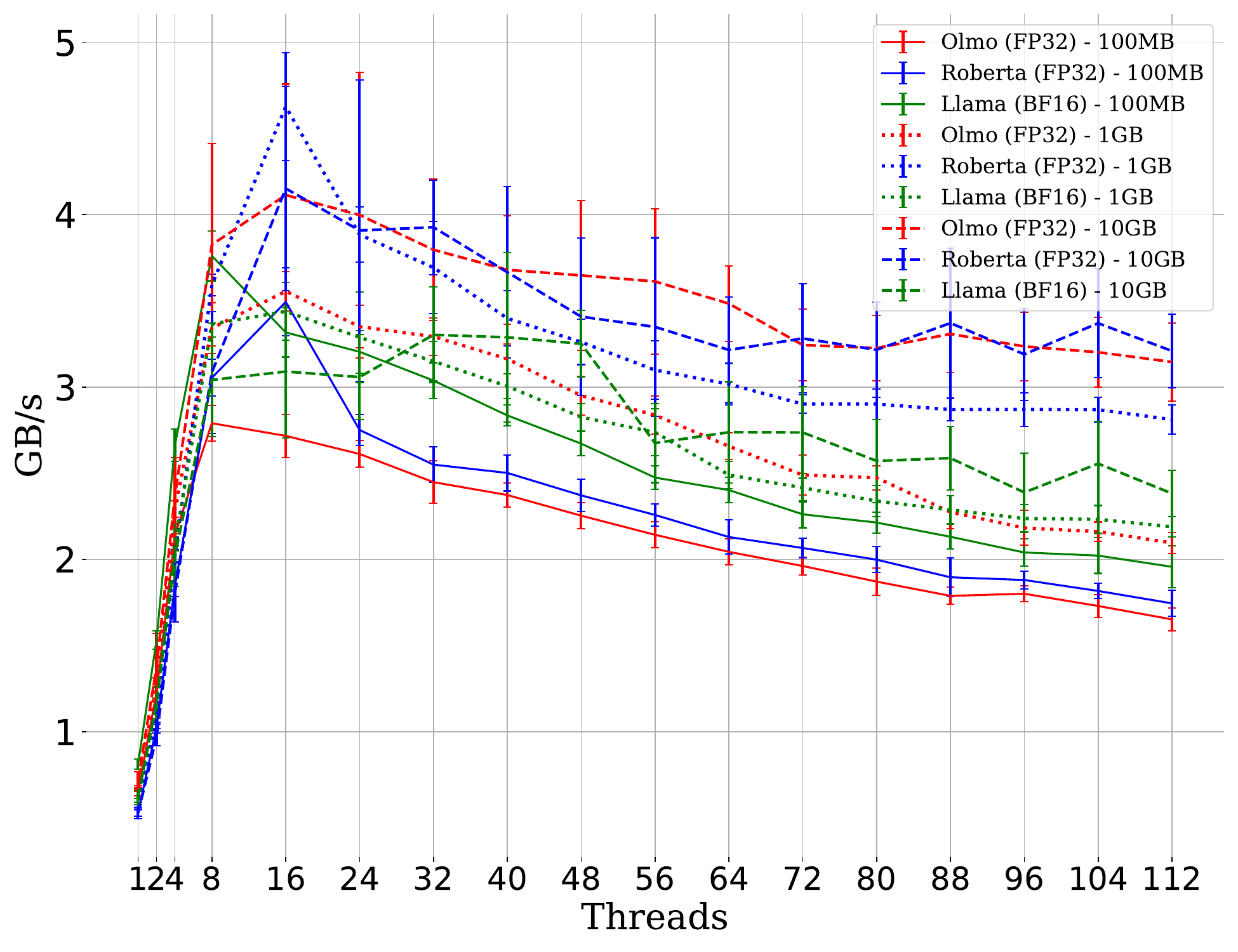}
        \vspace{-0.3in}
        \caption{Compression Throughput as a function of the number of threads used for compressing 100MB, 1GB, and 10GB across Olmo, Roberta, and Llama 3.1 models.}
        \label{fig:multithreadComp}
    \end{minipage} 
\end{figure}

\begin{figure}[ht]
%\vskip - 0.1in
    \begin{minipage}[b]{0.99\linewidth}
        \centering
        \includegraphics[width=\linewidth]{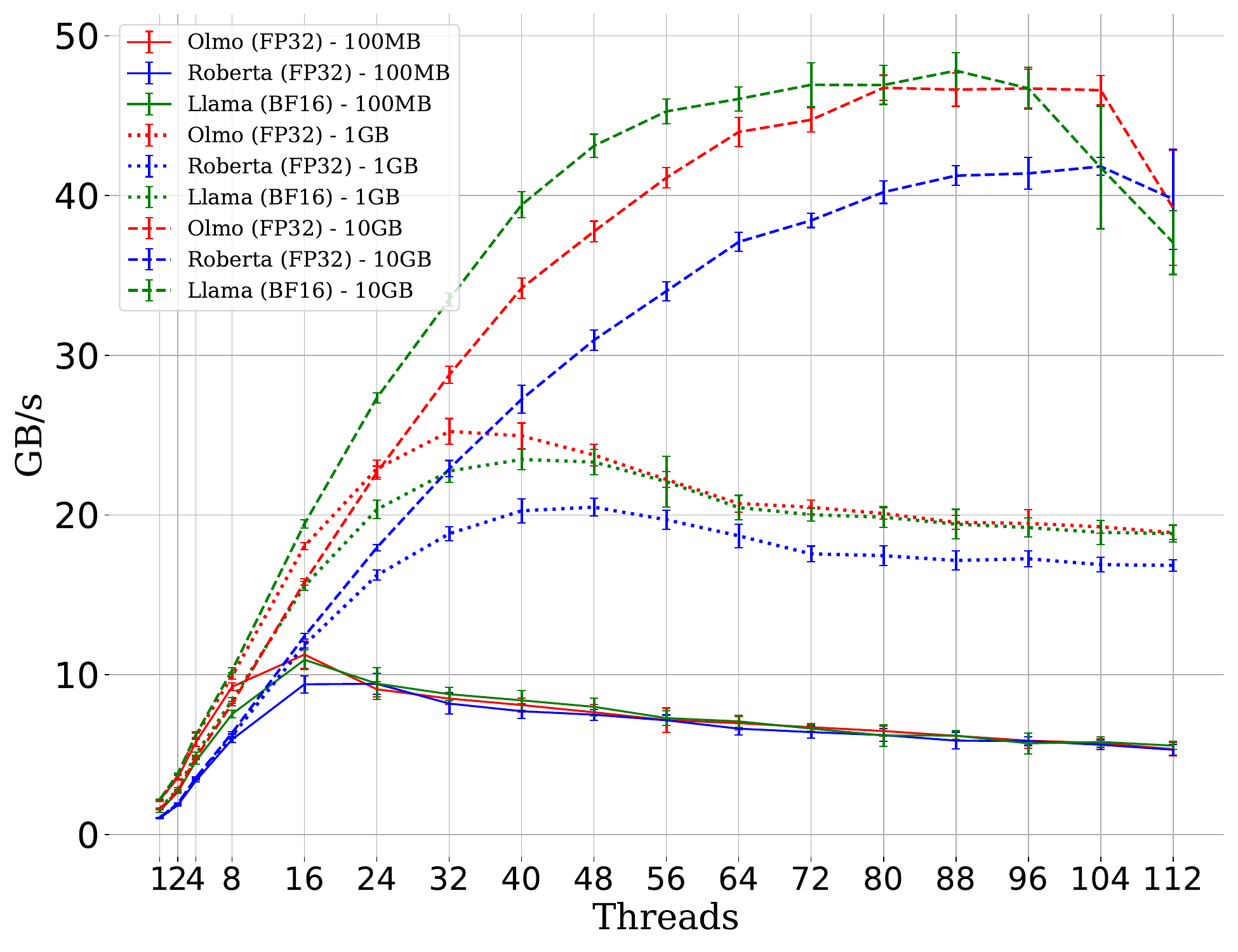}
        \vspace{-0.3in}
        \caption{Decompression Throughput as a function of the number of threads used for decompressing 100MB, 1GB, and 10GB across Olmo, Roberta, and Llama 3.1 models.}
        \label{fig:multithreadDecomp}
    \end{minipage} 
\end{figure}

\begin{figure}[ht]
%\vskip - 0.1in
    \begin{minipage}[b]{0.99\linewidth}
        \centering
        \includegraphics[width=\linewidth]{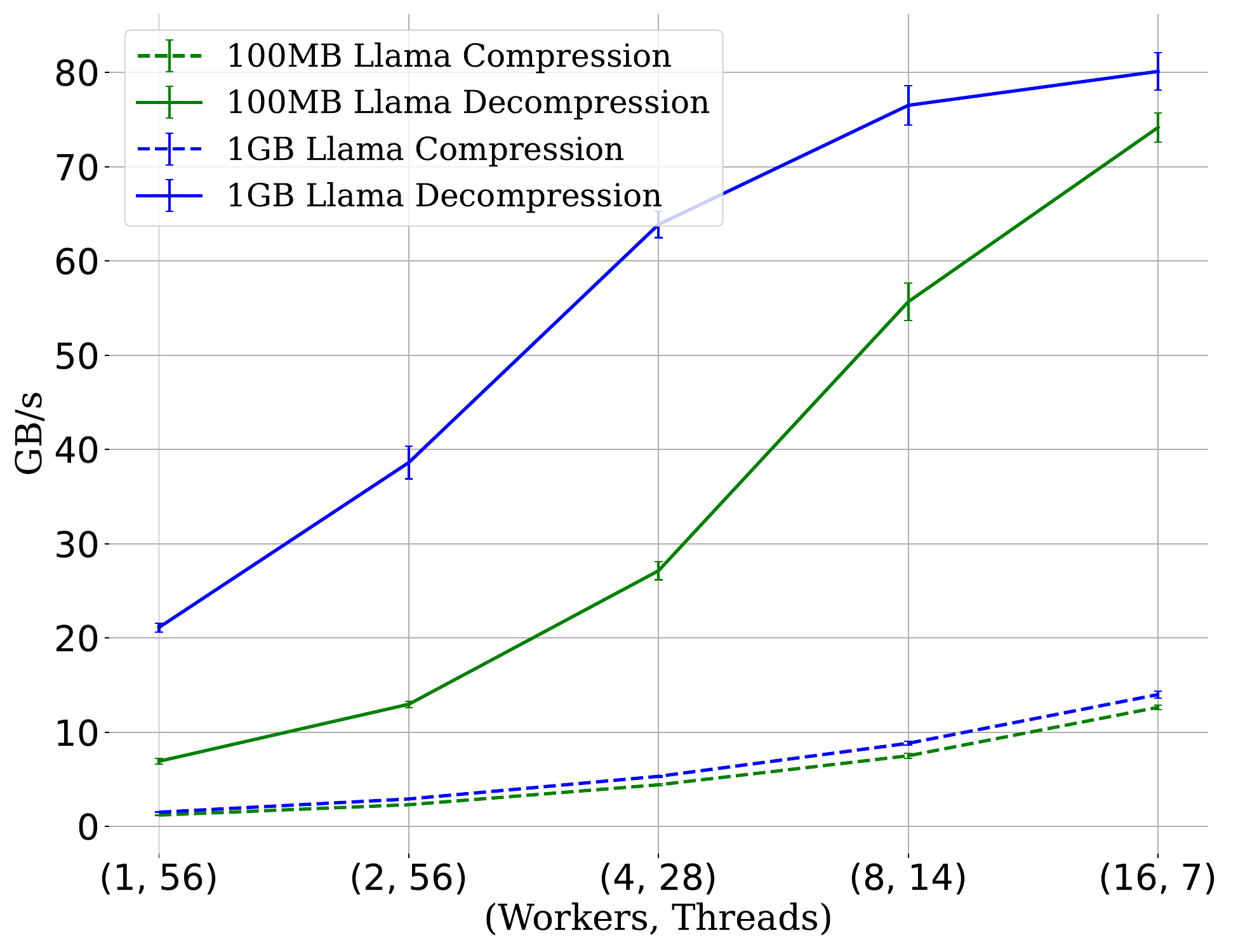}
        \vspace{-0.3in}
        \caption{Compression and Decompression Throughput of 100MB and 1GB of Llama 3.1 as a function of the number of workers and threads used.}
        \label{fig:workers}
    \end{minipage} 
\end{figure}

}

\subsection{End-2-End Evaluation}\label{sec:e2e}

\subsubsection{Model Hubs}\label{sec:HF}

In this section, we focus on time aspects and end-2-end timing of our first use-case - that of model hubs. 
We measured the time it takes to upload and download from Hugging~Face to a virtual machine that runs on one of the cloud providers and is located in the Milan region. We also measured upload and download performance on a home laptop with a 500Mbps network. 

Unlike storage benefits, communication speeds depend heavily on the medium. We first characterized the general behavior of the communication with the Hugging~Face hub. 
The upload bandwidth observed in the cloud remained mostly constant (at around 20 MBps).
For downloads, we observed 2 types of data transfer speeds. 

\noindent \textbf{First Download} - The speed in the first download showed large variance. On the cloud VM we measured 20-40 MBps. The home machine achieved approximately 10MBps.

\noindent \textbf{Cached Download} - The second read on the data is likely downloaded from a cloud cache and exhibits speed of 120-130 MBps on the cloud VM and approximately 40MBps at the home location. 

%\vskip -0.1in

\begin{figure}[ht]
%\vskip - 0.1in
    \begin{minipage}[b]{0.99\linewidth}
        \centering
        \includegraphics[width=\linewidth]{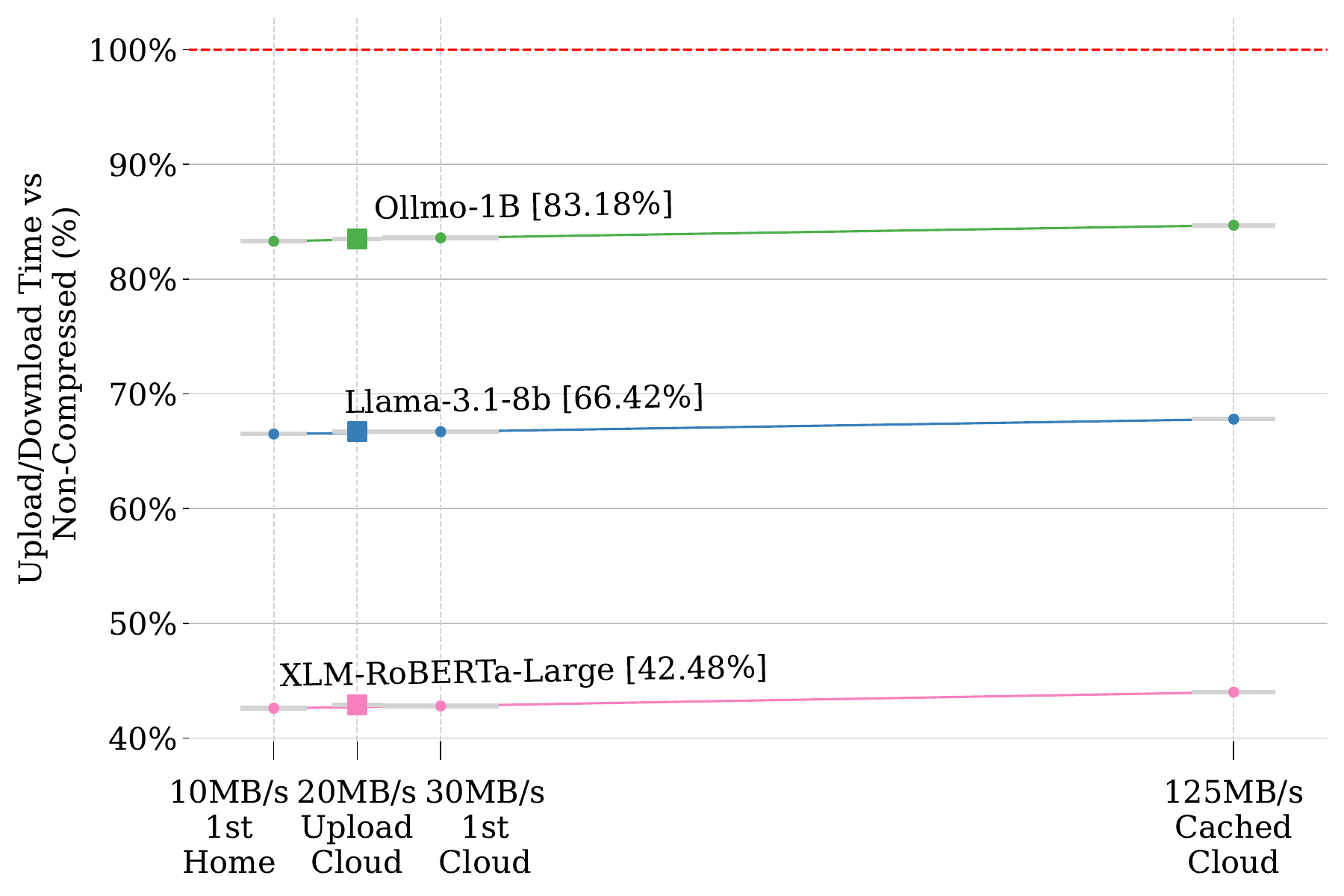}
        \vspace{-0.3in}
        \caption{Download and upload times of 3 models using our compression (with 8 threads) vs.\ the non-compressed version. }
        \label{fig:HFtimes}
    \end{minipage} 
\end{figure}

The end-2-end timing behavior is dictated by the time to decompress the model with 8 threads and the time to download it. 
We measured timing with our compression method on the 3 representative models tested in the previous section. Figure~\ref{fig:HFtimes} shows the timing of download speeds and also tests the effect on upload times to the cloud.
Each test was run 10 times for the cached reads and 5 times for the $1^{st}$ timers. The variance was almost entirely due to the network time and this standard deviation is depicted in the graph. The actual compression and decompression time had little variance. For example, for the xlm-RoBERTa model the average time for the decompression was 3.92 seconds with a standard deviation of 0.017. 

As expected, highly compressible models show significant time improvements whereas the less compressible models show more modest time savings. 
Naturally, the time saving is more significant when the network is slower. The upload time improves greatly since the bandwidth for uploads is low. On the other hand, the upload savings are lower than download with similar bandwidth reflecting the fact that compression is slower than decompression.  

\subsubsection{Inference Serving System}\label{sec:inference}
Another use case we evaluated is a high-end OpenShift cluster for inference serving.  
We measured model loading times in a cloned version of such a cluster using the same pod configuration described in section~\ref{sec:multithreading}.
The models are loaded from a high-performance clustered file system (having been previously downloaded from Hugging Face). The file system is connected to the pods via a Persistent Volume Claim (PVC) and exhibits a throughput of 8GB/s per worker. The test was run using vLLM version 0.7.2 with 4 workers, loading a 16GB Granite-3.1-8b-instruct BF16 model~\cite{granite2024granite} which had been compressed to $\frac{2}{3}$ of its original size.
Using our multi-threaded implementation, loading the compressed model from PVC to CPU and then to GPU (after decompression) took approximately 3 seconds, on par with running using a non-compressed model. This demonstrates that we can store models in compressed format on the PVC to save storage space with negligible impact on pod initialization time.

%\vspace{-0.1in}

\section {Related Work} \label{sec:related}
%While scaling models up \cite{Hoffmann2022TrainingCL,zhai2022scaling} and keeping multiple versions of them \cite{kandpal2023git,don-yehiya-etal-2023-cold} are commonly discussed, dealing with the size externalities is surprisingly not, except for inference size.
\subsection{Model Compression}
In the literature, "model-compression" is a field of its own, aimed at creating smaller models that mimic the original model. Model-compression is hence the name for a set of tools that aim to \emph{accelerate} models, usually at inference, by reducing their size \cite{choudhary2020comprehensive}. Under such conditions, a method is allowed to reduce the accuracy, and is judged on its tradeoff between size and performance. This differs from lossless compression which is supposed to return the model to its original state after decompression. 

There are four main methods to reduce model size in that manner \cite{choudhary2020comprehensive}. Pruning \cite{LeCun1989OptimalBD,Hanson1988ComparingBF,zhu2017prune} (sometimes referred to as sparsification;  \cite{ma2021effective}) where parts of the model are removed, dedicated training or network architecture \cite{oktay2019scalable}, distillation \cite{gou2021knowledge} or otherwise training a smaller model from a better model \cite{Haroush2019TheKW}, and quantization \cite{gholami2021survey_quantization}. There are also methods that combine several of those \cite{polino2018model, Han2015DeepCC}.

%\subsection{Quantization}
Out of the model-compression techniques, quantization  \cite{gray1984vector} is perhaps the most popular. Quantization is a method that bins weight values to a more coarse granularity. By design it trades-off accuracy for space and speed and is hence a lossy compression technique. Quantization is bound to increase the amount of entropy per byte in a model but is not trying to push it to the limit. As such, quantized models can potentially be further compressed using lossless compression. We see mixed results: with some off-the-shelf quantized models that have been quantized with GTPQ~\cite{frantar2022gptq} and AWQ~\cite{lin2023awq} we find that they are still compressible, with a compressed size between 85-91\%. On the other hand, we found that other quantized models using GGUF do not compress at all. 

%In a sense, quantization is complementary to compression. Quantization is able to improve inference speed and to reduce model size  drastically, at costs to inference accuracy or the ability to further train the model. In contrast, compression cannot speed inference but can compress models further, or compress a model without affecting its behaviour at all. Similar to quantization, our Tunable lossy compression drops some of the information, but only ones that don't reduce accuracy in a measurable way. Unlike quantization, it changes the representation which is expected to further reduce the size as seen by the results above. % as discussed in Section~\ref{sec:compress_quantized}). %For example, quantization requires saving in floating point format, as it will be required for fast multiplication during inference. 

% original Quantization thoughts:
% It is about the combination of transforming the values and feeding them into a lossless compressor. 
% This is different than standard quantization as in our patent the transformation part does not map the values into a smaller range but rather keeps the original number of bits in the parameter. The reduction in size is instead achieved by the lossless compressor. This method achieves flexibility of the precision factor that is not possible (or counterproductive) when doing quantization.

% Add here all the papers we shared between us in the channel/DMs

\subsection{Floating Point Compression}

There is a body of literature on compression tailored specifically for floating-point data type. 
%Chen et el. ~\cite{} proposed separating the "sign exponent" from the mantissa and compressing it independently with gzip. This approach was part of a more complex method that involved among others compressing differences in the mantissa. 
% MPC: also suggest compressing "sign exponent" but only as part of a 4 part of compression method.
Some compression methods~\cite{chen2006lossless, yang2015mpc} propose separating the
"sign exponent" from the mantissa and compressing it independently as part of a more complex approach.
%In \zipnn{}, we compress the exponent using entropy encoding without searching for multi-byte repetitions and do not compress the sign bit or mantissa unless necessary.

%\begin{comment}
%zfp algorithm:
% https://zfp.readthedocs.io/en/latest/algorithm.html#algorithm
%\end{comment}

% However, to the best of our knowledge, no prior work has investigated exponent extraction compression~\ref{sec:model_compressibility} across diverse AI models. Our approach shows that compressing only the exponent with simple entropy encoding, while leaving the sign bit and mantissa uncompressed, achieves high compression ratios across different model architectures.
\textit{dietgpu}~\cite{dietgpu} is an open source project that implements ANS compression in the GPU aiming to reduce incoming traffic to a GPU. One of their variants implements exponent only compression for floating point streams is similar to \zipnn{} on regular models. This falls short on clean models, checkpoint deltas, gradients and optimizers. 

Other compression methods, such as ZFP~\cite{zfpgithub}, based on~\cite{diffenderfer2019error, lindstrom2006fast}, take a block-based approach by dividing data into small blocks. Within each block, floating-point values are converted to a block-floating-point format, sharing a single exponent while being transformed into signed integers.
Compression is then applied to this representation, making it particularly effective for lossy compression with controllable error bounds.
Our benchmarks using ZFP version 1.3.0 with FP32 show its compression ratio comparable to ZSTD but lower than \zipnn{}'s compression ratio.

% Add a lengthy discussion about the difference from quantization!
\subsection{Other Related Work}
%Interestingly, two recent works also found improved results by reducing some of the information in the network (see \S\ref{sec:losseless_res}). They saw it during pruning \cite{Sharma2023TheTI} or pruning and extreme quantization \cite{yadav2023compeft}. We observe this same phenomena in the tunable lossy compression. Future work may find a theory to connect those findings.

Similar to delta compression, some works analyze dimensionality \cite{aghajanyan-etal-2021-intrinsic, gueta-etal-2023-knowledge} or save the deltas for actions such as compositionally \cite{ilharco2022editing} and merging multiple deltas \cite{Choshen2022FusingFM,Wortsman2022ModelSA,matena2021merging}. Few works also apply on such deltas methods like pruning \cite{yadav2023ties}, trained sparsity \cite{zhang2023adaptive} or quantization \cite{dettmers2023qlora}.

Another line of work worth mentioning is computation graph optimization \cite{sabne2020xla,wu2023pytorch}. Such works reduce the computation graph and perform optimization there. It is noteworthy to contrast it to our work. such work compresses the size of the computation graph, which speeds computation, but does not change the model weights, and it is hence orthogonal to our work. To validate that, we compressed pyTorch \cite{wu2023pytorch} models before and after compilation, finding compression works similarly well.
Last, related to the optimization process (see \S\ref{sec:gradients}), a few recent works offer to reduce the information passed in gradient updates hence making them faster, further overcoming the decrease in information per example by seeing more examples per second \cite{tyagi2023gravac,zhao2024galore}.

%Independent of our work, we discovered a GitHub repository called \textit{dietgpu}~\cite{dietgpu} focused on compressing AI models on the GPU. They have an extension for Adaptive Numerical Systems (ANS) compression on the GPU. They have an extension for compression and decompression of unstructured floating point data, which, according to our benchmarks, achieves a compression ratio closer to \zipnn{}. The authors do not elaborate on their methodology, but by examining the code they use a similar approach to exponent extraction~\ref{sec:model_compressibility}. In addition, the authors have not published any compression ratio results

%From DietGPU: An extension to the above to handle fast lossless compression and decompression of unstructured floating point data, for use in ML and HPC applications, especially in communicating over local interconnects (PCIe / NVLink) and remote interconnects (Ethernet / InfiniBand). This operates at around 250-600 GB/s for reasonable data sizes on an A100 GPU.

\section {Conclusion}
We are in an era where models and system requirements grow larger -- overparametrization seems to be beneficial for better learning. As our compression findings hint, this overparametrization is not fully used for inference or for the weights themselves and there is redundancy. Hence the wide attention and progress made to reducing model sizes is not without merit. 
That being said, the reality is that commonly used models are not kept or run in reduced form and there is great inefficiency in the way models are stored and communicated today. Some of this inefficiency can be mitigated using the compression techniques outlined in this paper. Even more so, by tailoring our compression to models, one can achieve significantly better compression with significantly smaller overhead. 

Given the reduction in network bandwidth, storage, and time, we think that lossless compression should be the default in communication with model hubs such as Hugging~Face. Moreover, we believe that communication compression has multiple other use-cases in the realm of training, versioning, and serving models. 

%\paragraph{Impact Statement:} This paper presents work whose goal is to advance the field of Machine Learning. The improvements we offer are in the area of making models more accessible and easing collaboration. Our work is orthogonal to the abilities and societal issues that AI models may raise. As such, we do not see any potential societal consequences of our work that must be specifically highlighted here.

%\moshik{add the fact that the deocmpression throughput can reach up to 80GB/s}

%\begin{thebibliography}{00}
\bibliography{compression, bib}

\begin{thebibliography}{10}

\bibitem{Biderman2023PythiaAS}
S.~Biderman, H.~Schoelkopf, Q.~G. Anthony, H.~Bradley, K.~O'Brien, E.~Hallahan, M.~A. Khan, S.~Purohit, U.~S. Prashanth, E.~Raff, A.~Skowron, L.~Sutawika, and O.~van~der Wal, ``Pythia: A suite for analyzing large language models across training and scaling,'' {\em ArXiv}, vol.~abs/2304.01373, 2023.

\bibitem{kandpal2023git}
N.~Kandpal, B.~Lester, M.~Muqeeth, A.~Mascarenhas, M.~Evans, V.~Baskaran, T.~Huang, H.~Liu, and C.~Raffel, ``Git-theta: A git extension for collaborative development of machine learning models,'' {\em arXiv preprint arXiv:2306.04529}, 2023.

\bibitem{don-yehiya-etal-2023-cold}
S.~Don-Yehiya, E.~Venezian, C.~Raffel, N.~Slonim, and L.~Choshen, ``{C}ol{D} fusion: Collaborative descent for distributed multitask finetuning,'' in {\em Proceedings of the 61st Annual Meeting of the Association for Computational Linguistics (Volume 1: Long Papers)} (A.~Rogers, J.~Boyd-Graber, and N.~Okazaki, eds.), (Toronto, Canada), pp.~788--806, Association for Computational Linguistics, July 2023.

\bibitem{zhang2021survey}
C.~Zhang, Y.~Xie, H.~Bai, B.~Yu, W.~Li, and Y.~Gao, ``A survey on federated learning,'' {\em Knowledge-Based Systems}, vol.~216, p.~106775, 2021.

\bibitem{Wolf2019HuggingFacesTS}
T.~Wolf, L.~Debut, V.~Sanh, J.~Chaumond, C.~Delangue, A.~Moi, P.~Cistac, T.~Rault, R.~Louf, M.~Funtowicz, and J.~Brew, ``Huggingface's transformers: State-of-the-art natural language processing,'' {\em ArXiv}, vol.~abs/1910.03771, 2019.

\bibitem{jiang2023mistral}
A.~Q. Jiang, A.~Sablayrolles, A.~Mensch, C.~Bamford, D.~S. Chaplot, D.~d.~l. Casas, F.~Bressand, G.~Lengyel, G.~Lample, L.~Saulnier, {\em et~al.}, ``Mistral 7b,'' {\em arXiv preprint arXiv:2310.06825}, 2023.

\bibitem{gou2021knowledge}
J.~Gou, B.~Yu, S.~J. Maybank, and D.~Tao, ``Knowledge distillation: A survey,'' {\em International Journal of Computer Vision}, vol.~129, pp.~1789--1819, 2021.

\bibitem{ma2021effective}
X.~Ma, M.~Qin, F.~Sun, Z.~Hou, K.~Yuan, Y.~Xu, Y.~Wang, Y.-K. Chen, R.~Jin, and Y.~Xie, ``Effective model sparsification by scheduled grow-and-prune methods,'' {\em arXiv preprint arXiv:2106.09857}, 2021.

\bibitem{gholami2021survey_quantization}
A.~Gholami, S.~Kim, Z.~Dong, Z.~Yao, M.~W. Mahoney, and K.~Keutzer, ``A survey of quantization methods for efficient neural network inference,'' 2021.

\bibitem{deutsch1996zlib}
P.~Deutsch and J.-L. Gailly, ``Zlib compressed data format specification version 3.3,'' tech. rep., 1996.

\bibitem{collet2018zstandard}
Y.~Collet and M.~Kucherawy, ``Zstandard compression and the application/zstd media type,'' tech. rep., 2018.

\bibitem{lZ77}
J.~Ziv and A.~Lempel, ``A universal algorithm for sequential data compression,'' {\em IEEE Transactions on Information Theory}, vol.~23, no.~3, pp.~337--343, 1977.

\bibitem{LZ78}
J.~Ziv and A.~Lempel, ``{Compression of Individual Sequences via Variable-Rate Coding},'' {\em {IEEE Transactions on Information Theory}}, vol.~{24}, pp.~530--536, {September} 1978.

\bibitem{Huffman1952}
D.~A. Huffman, ``A method for the construction of minimum-redundancy codes,'' {\em Proceedings of the IRE}, vol.~40, no.~9, pp.~1098--1101, 1952.

\bibitem{liu2019roberta}
Y.~Liu, M.~Ott, N.~Goyal, J.~Du, M.~Joshi, D.~Chen, O.~Levy, M.~Lewis, L.~Zettlemoyer, and V.~Stoyanov, ``Roberta: A robustly optimized bert pretraining approach,'' {\em arXiv preprint arXiv:1907.11692}, 2019.

\bibitem{Devlin2019BERTPO}
J.~Devlin, M.-W. Chang, K.~Lee, and K.~Toutanova, ``Bert: Pre-training of deep bidirectional transformers for language understanding,'' in {\em North American Chapter of the Association for Computational Linguistics}, 2019.

\bibitem{Fedus2021SwitchTS}
W.~Fedus, B.~Zoph, and N.~M. Shazeer, ``Switch transformers: Scaling to trillion parameter models with simple and efficient sparsity,'' {\em J. Mach. Learn. Res.}, vol.~23, pp.~120:1--120:39, 2021.

\bibitem{modelZoo}
J.~Yu~Koh, ``Model zoo (hub),'' 2018.

\bibitem{Pytorch_2019}
Pytorch, ``Pytorch hub,'' 2019.

\bibitem{tensorflowHub}
Google, ``Tensorflow hub,'' 2018.

\bibitem{adapterhub}
J.~Pfeiffer, A.~R{\"u}ckl{\'e}, C.~Poth, A.~Kamath, I.~Vuli{\'c}, S.~Ruder, K.~Cho, and I.~Gurevych, ``{A}dapter{H}ub: A framework for adapting transformers,'' in {\em Proceedings of the 2020 Conference on Empirical Methods in Natural Language Processing: System Demonstrations} (Q.~Liu and D.~Schlangen, eds.), (Online), pp.~46--54, Association for Computational Linguistics, Oct. 2020.

\bibitem{watsonxdata}
``Ibm watsonx.data.'' https://www.ibm.com/products/watsonx-data.

\bibitem{QualcommAIHUB}
``Qualcomm® ai hub.'' https://aihub.qualcomm.com/.

\bibitem{zhao2023pytorch}
Y.~Zhao, A.~Gu, R.~Varma, L.~Luo, C.-C. Huang, M.~Xu, L.~Wright, H.~Shojanazeri, M.~Ott, S.~Shleifer, {\em et~al.}, ``Pytorch fsdp: experiences on scaling fully sharded data parallel,'' {\em arXiv preprint arXiv:2304.11277}, 2023.

\bibitem{li2022branch}
M.~Li, S.~Gururangan, T.~Dettmers, M.~Lewis, T.~Althoff, N.~A. Smith, and L.~Zettlemoyer, ``Branch-train-merge: Embarrassingly parallel training of expert language models,'' {\em arXiv preprint arXiv:2208.03306}, 2022.

\bibitem{lialin2023relora}
V.~Lialin, S.~Muckatira, N.~Shivagunde, and A.~Rumshisky, ``Relora: High-rank training through low-rank updates,'' in {\em Workshop on Advancing Neural Network Training: Computational Efficiency, Scalability, and Resource Optimization (WANT@ NeurIPS 2023)}, 2023.

\bibitem{NEURIPS2021_41a60377}
M.~Diskin, A.~Bukhtiyarov, M.~Ryabinin, L.~Saulnier, q.~lhoest, A.~Sinitsin, D.~Popov, D.~V. Pyrkin, M.~Kashirin, A.~Borzunov, A.~Villanova~del Moral, D.~Mazur, I.~Kobelev, Y.~Jernite, T.~Wolf, and G.~Pekhimenko, ``Distributed deep learning in open collaborations,'' in {\em Advances in Neural Information Processing Systems} (M.~Ranzato, A.~Beygelzimer, Y.~Dauphin, P.~Liang, and J.~W. Vaughan, eds.), vol.~34, pp.~7879--7897, Curran Associates, Inc., 2021.

\bibitem{Turner2021BayesianOI}
R.~Turner, D.~Eriksson, M.~J. McCourt, J.~Kiili, E.~Laaksonen, Z.~Xu, and I.~M. Guyon, ``Bayesian optimization is superior to random search for machine learning hyperparameter tuning: Analysis of the black-box optimization challenge 2020,'' in {\em Neural Information Processing Systems}, 2021.

\bibitem{Dodge2020FineTuningPL}
J.~Dodge, G.~Ilharco, R.~Schwartz, A.~Farhadi, H.~Hajishirzi, and N.~A. Smith, ``Fine-tuning pretrained language models: Weight initializations, data orders, and early stopping,'' {\em ArXiv}, vol.~abs/2002.06305, 2020.

\bibitem{JunczysDowmunt2018MarianFN}
M.~Junczys-Dowmunt, R.~Grundkiewicz, T.~Dwojak, H.~T. Hoang, K.~Heafield, T.~Neckermann, F.~Seide, U.~Germann, A.~F. Aji, N.~Bogoychev, A.~F.~T. Martins, and A.~Birch, ``Marian: Fast neural machine translation in c++,'' in {\em Annual Meeting of the Association for Computational Linguistics}, 2018.

\bibitem{Sandler2023TrainingTM}
M.~Sandler, A.~Zhmoginov, M.~Vladymyrov, and N.~Miller, ``Training trajectories, mini-batch losses and the curious role of the learning rate,'' {\em ArXiv}, vol.~abs/2301.02312, 2023.

\bibitem{liu2023llm360}
Z.~Liu, A.~Qiao, W.~Neiswanger, H.~Wang, B.~Tan, T.~Tao, J.~Li, Y.~Wang, S.~Sun, O.~Pangarkar, R.~Fan, Y.~Gu, V.~Miller, Y.~Zhuang, G.~He, H.~Li, F.~Koto, L.~Tang, N.~Ranjan, Z.~Shen, X.~Ren, R.~Iriondo, C.~Mu, Z.~Hu, M.~Schulze, P.~Nakov, T.~Baldwin, and E.~P. Xing, ``Llm360: Towards fully transparent open-source llms,'' {\em arXiv}, 2023.

\bibitem{shoeybi2019megatron}
M.~Shoeybi, M.~Patwary, R.~Puri, P.~LeGresley, J.~Casper, and B.~Catanzaro, ``Megatron-lm: Training multi-billion parameter language models using model parallelism,'' {\em arXiv preprint arXiv:1909.08053}, 2019.

\bibitem{granite2024granite}
I.~Granite~Team, ``Granite 3.0 language models,'' 2024.

\bibitem{dubey2024llama}
A.~Dubey, A.~Jauhri, A.~Pandey, A.~Kadian, A.~Al-Dahle, A.~Letman, A.~Mathur, A.~Schelten, A.~Yang, A.~Fan, {\em et~al.}, ``The llama 3 herd of models,'' {\em arXiv preprint arXiv:2407.21783}, 2024.

\bibitem{wang2019bfloat16}
S.~Wang and P.~Kanwar, ``Bfloat16: The secret to high performance on cloud tpus,'' {\em Google Cloud Blog}, vol.~4, 2019.

\bibitem{squash}
Squash, ``Squash compression benchmark,'' 2016.

\bibitem{Riss1976AC}
J.~J. Rissanen, ``Generalized kraft inequality and arithmetic coding,'' {\em IBM Journal of Research and Development}, vol.~20, no.~3, pp.~198--203, 1976.

\bibitem{zlib}
M.~Adler and J.-L. Gailly, ``Zlib,'' 2024.

\bibitem{zstd}
Y.~Collet, ``Zstandard,'' 2024.

\bibitem{lz4}
Y.~Collet, ``Lz4 - extremely fast compression,'' 2024.

\bibitem{snappy}
J.~Dean, S.~Ghemawat, and S.~H. Gunderson, ``Snappy, a fast compressor/decompressor.,'' 2024.

\bibitem{tozip2013}
D.~Harnik, R.~Kat, D.~Sotnikov, A.~Traeger, and O.~Margalit, ``To zip or not to zip: Effective resource usage for {Real-Time} compression,'' in {\em 11th USENIX Conference on File and Storage Technologies (FAST 13)}, (San Jose, CA), pp.~229--241, USENIX Association, Feb. 2013.

\bibitem{anil2020scalable}
R.~Anil, V.~Gupta, T.~Koren, K.~Regan, and Y.~Singer, ``Scalable second order optimization for deep learning,'' {\em arXiv preprint arXiv:2002.09018}, 2020.

\bibitem{ZipNNRepo}
``Zipnn: A lossless compression library for ai pipelines.'' https://github.com/zipnn/zipnn, 2024.

\bibitem{wolf-etal-2020-transformers}
T.~Wolf, L.~Debut, V.~Sanh, J.~Chaumond, C.~Delangue, A.~Moi, P.~Cistac, T.~Rault, R.~Louf, M.~Funtowicz, J.~Davison, S.~Shleifer, P.~von Platen, C.~Ma, Y.~Jernite, J.~Plu, C.~Xu, T.~Le~Scao, S.~Gugger, M.~Drame, Q.~Lhoest, and A.~Rush, ``Transformers: State-of-the-art natural language processing,'' in {\em Proceedings of the 2020 Conference on Empirical Methods in Natural Language Processing: System Demonstrations} (Q.~Liu and D.~Schlangen, eds.), (Online), pp.~38--45, Association for Computational Linguistics, Oct. 2020.

\bibitem{choudhary2020comprehensive}
T.~Choudhary, V.~Mishra, A.~Goswami, and J.~Sarangapani, ``A comprehensive survey on model compression and acceleration,'' {\em Artificial Intelligence Review}, vol.~53, pp.~5113--5155, 2020.

\bibitem{LeCun1989OptimalBD}
Y.~LeCun, J.~S. Denker, and S.~A. Solla, ``Optimal brain damage,'' in {\em Neural Information Processing Systems}, 1989.

\bibitem{Hanson1988ComparingBF}
S.~J. Hanson and L.~Y. Pratt, ``Comparing biases for minimal network construction with back-propagation,'' in {\em Neural Information Processing Systems}, 1988.

\bibitem{zhu2017prune}
M.~Zhu and S.~Gupta, ``To prune, or not to prune: exploring the efficacy of pruning for model compression,'' {\em arXiv preprint arXiv:1710.01878}, 2017.

\bibitem{oktay2019scalable}
D.~Oktay, J.~Ball{\'e}, S.~Singh, and A.~Shrivastava, ``Scalable model compression by entropy penalized reparameterization,'' {\em arXiv preprint arXiv:1906.06624}, 2019.

\bibitem{Haroush2019TheKW}
M.~Haroush, I.~Hubara, E.~Hoffer, and D.~Soudry, ``The knowledge within: Methods for data-free model compression,'' {\em 2020 IEEE/CVF Conference on Computer Vision and Pattern Recognition (CVPR)}, pp.~8491--8499, 2019.

\bibitem{polino2018model}
A.~Polino, R.~Pascanu, and D.~Alistarh, ``Model compression via distillation and quantization,'' {\em arXiv preprint arXiv:1802.05668}, 2018.

\bibitem{Han2015DeepCC}
S.~Han, H.~Mao, and W.~J. Dally, ``Deep compression: Compressing deep neural network with pruning, trained quantization and huffman coding,'' {\em arXiv: Computer Vision and Pattern Recognition}, 2015.

\bibitem{gray1984vector}
R.~Gray, ``Vector quantization,'' {\em IEEE Assp Magazine}, vol.~1, no.~2, pp.~4--29, 1984.

\bibitem{frantar2022gptq}
E.~Frantar, S.~Ashkboos, T.~Hoefler, and D.~Alistarh, ``Gptq: Accurate post-training quantization for generative pre-trained transformers,'' {\em arXiv preprint arXiv:2210.17323}, 2022.

\bibitem{lin2023awq}
J.~Lin, J.~Tang, H.~Tang, S.~Yang, X.~Dang, and S.~Han, ``Awq: Activation-aware weight quantization for llm compression and acceleration,'' {\em arXiv preprint arXiv:2306.00978}, 2023.

\bibitem{chen2006lossless}
D.~Chen, Y.-J. Chiang, N.~D. Memon, and X.~Wu, ``Lossless geometry compression for steady-state and time-varying irregular grids.,'' in {\em EuroVis}, pp.~275--282, 2006.

\bibitem{yang2015mpc}
A.~Yang, H.~Mukka, F.~Hesaaraki, and M.~Burtscher, ``Mpc: a massively parallel compression algorithm for scientific data,'' in {\em 2015 IEEE International Conference on Cluster Computing}, pp.~381--389, IEEE, 2015.

\bibitem{dietgpu}
M.~Levental, J.~Johnson, {\em et~al.}, ``dietgpu.''

\bibitem{zfpgithub}
P.~Lindstrom {\em et~al.}, ``{ZFP}.''

\bibitem{diffenderfer2019error}
J.~Diffenderfer, A.~Fox, J.~A.~F. Hittinger, G.~Sanders, and P.~G. Lindstrom, ``Error analysis of {ZFP} compression for floating-point data,'' {\em {SIAM} J. Sci. Comput.}, vol.~41, no.~3, pp.~A1867--A1898, 2019.

\bibitem{lindstrom2006fast}
P.~Lindstrom and M.~Isenburg, ``Fast and efficient compression of floating-point data,'' {\em IEEE transactions on visualization and computer graphics}, vol.~12, no.~5, pp.~1245--1250, 2006.

\bibitem{aghajanyan-etal-2021-intrinsic}
A.~Aghajanyan, S.~Gupta, and L.~Zettlemoyer, ``Intrinsic dimensionality explains the effectiveness of language model fine-tuning,'' in {\em Proceedings of the 59th Annual Meeting of the Association for Computational Linguistics and the 11th International Joint Conference on Natural Language Processing (Volume 1: Long Papers)} (C.~Zong, F.~Xia, W.~Li, and R.~Navigli, eds.), (Online), pp.~7319--7328, Association for Computational Linguistics, Aug. 2021.

\bibitem{gueta-etal-2023-knowledge}
A.~Gueta, E.~Venezian, C.~Raffel, N.~Slonim, Y.~Katz, and L.~Choshen, ``Knowledge is a region in weight space for fine-tuned language models,'' in {\em Findings of the Association for Computational Linguistics: EMNLP 2023} (H.~Bouamor, J.~Pino, and K.~Bali, eds.), (Singapore), pp.~1350--1370, Association for Computational Linguistics, Dec. 2023.

\bibitem{ilharco2022editing}
G.~Ilharco, M.~T. Ribeiro, M.~Wortsman, S.~Gururangan, L.~Schmidt, H.~Hajishirzi, and A.~Farhadi, ``Editing models with task arithmetic,'' {\em arXiv preprint arXiv:2212.04089}, 2022.

\bibitem{Choshen2022FusingFM}
L.~Choshen, E.~Venezian, N.~Slonim, and Y.~Katz, ``Fusing finetuned models for better pretraining,'' {\em ArXiv}, vol.~abs/2204.03044, 2022.

\bibitem{Wortsman2022ModelSA}
M.~Wortsman, G.~Ilharco, S.~Y. Gadre, R.~Roelofs, R.~Gontijo-Lopes, A.~S. Morcos, H.~Namkoong, A.~Farhadi, Y.~Carmon, S.~Kornblith, and L.~Schmidt, ``Model soups: averaging weights of multiple fine-tuned models improves accuracy without increasing inference time,'' 2022.

\bibitem{matena2021merging}
M.~Matena and C.~Raffel, ``Merging models with fisher-weighted averaging,'' {\em arXiv preprint arXiv:2111.09832}, 2021.

\bibitem{yadav2023ties}
P.~Yadav, D.~Tam, L.~Choshen, C.~Raffel, and M.~Bansal, ``Ties-merging: Resolving interference when merging models,'' in {\em Thirty-seventh Conference on Neural Information Processing Systems}, 2023.

\bibitem{zhang2023adaptive}
Q.~Zhang, M.~Chen, A.~Bukharin, P.~He, Y.~Cheng, W.~Chen, and T.~Zhao, ``Adaptive budget allocation for parameter-efficient fine-tuning,'' {\em arXiv preprint arXiv:2303.10512}, 2023.

\bibitem{dettmers2023qlora}
T.~Dettmers, A.~Pagnoni, A.~Holtzman, and L.~Zettlemoyer, ``Qlora: Efficient finetuning of quantized llms,'' {\em arXiv preprint arXiv:2305.14314}, 2023.

\bibitem{sabne2020xla}
A.~Sabne, ``Xla: Compiling machine learning for peak performance,'' 2020.

\bibitem{wu2023pytorch}
P.~Wu, ``Pytorch 2.0: The journey to bringing compiler technologies to the core of pytorch (keynote),'' in {\em Proceedings of the 21st ACM/IEEE International Symposium on Code Generation and Optimization}, pp.~1--1, 2023.

\bibitem{tyagi2023gravac}
S.~Tyagi and M.~Swany, ``Gravac: Adaptive compression for communication-efficient distributed dl training,'' in {\em 2023 IEEE 16th International Conference on Cloud Computing (CLOUD)}, pp.~319--329, IEEE, 2023.

\bibitem{zhao2024galore}
J.~Zhao, Z.~Zhang, B.~Chen, Z.~Wang, A.~Anandkumar, and Y.~Tian, ``Galore: Memory-efficient llm training by gradient low-rank projection,'' {\em arXiv preprint arXiv:2403.03507}, 2024.

\end{thebibliography}
\bibliographystyle{ieeetr}

\end{document}